\documentclass[letterpaper]{article} 
\usepackage{aaai2027}  
\usepackage{times}  
\usepackage{helvet}  
\usepackage{courier}  
\usepackage[hyphens]{url}  
\usepackage{graphicx} 
\usepackage{natbib}  
\usepackage{caption} 
\usepackage{microtype}
\usepackage{inconsolata}
\usepackage{url}            
\usepackage{subcaption}
\usepackage{booktabs}       
\usepackage{amsfonts}       
\usepackage{nicefrac}       
\usepackage{amsfonts}       
\usepackage{amsmath}        
\usepackage{enumitem}
\usepackage{makecell}
\usepackage{graphicx}
\usepackage{multirow}
\usepackage{subfig}
\usepackage{amsmath}
\usepackage{amssymb}

\usepackage[table]{xcolor}
\usepackage{float}
\usepackage{rotating}
\usepackage{arydshln}
\usepackage{amsthm}
\usepackage{tabularx}
\usepackage{array}
\usepackage[ruled,vlined]{algorithm2e}

\usepackage{newfloat}
\usepackage{listings}
\DeclareCaptionStyle{ruled}{labelfont=normalfont,labelsep=colon,strut=off} 
\floatstyle{ruled}
\newfloat{listing}{tb}{lst}{}
\floatname{listing}{Listing}
\title{Bidirectional Centered Score Matching for Discrete Diffusion Distillation}
\author{
    \large {Peng Tu\thanks{Equal Contribution}\textsuperscript{\rm 1}},
    \large {Baoyou Chen\footnotemark[1]\textsuperscript{\rm 1, \rm 3}},
    \large {Haoyuan Xia\textsuperscript{\rm 2}},
    \large {Zilong Dong\textsuperscript{\rm 4}},
    \large {Weihao Yuan\textsuperscript{\rm 5}}, 
    \large {Siyu Zhu\thanks{Corresponding Author}\textsuperscript{\rm 1, \rm 2, \rm 3}}
}
\affiliations{
    \textsuperscript{\rm 1}Shanghai Academy of AI for Science, 
    \textsuperscript{\rm 2}Shanghai Innovation Institute, \\
    \textsuperscript{\rm 3}Fudan University, 
    \textsuperscript{\rm 4}Alibaba Group,
    \textsuperscript{\rm 5}Nanjing University
}

\usepackage{bibentry}
\begin{document}

\maketitle

 \begin{abstract}
Discrete diffusion models with parallel decoding are often
distilled using token-wise KL objectives. Although token-wise
KL exactly matches the corresponding product-form block
distributions, its probability-space optimization geometry and
teacher-only candidate truncation may provide insufficient
supervision under aggressive few-step distillation. Concrete
Score Matching (CSM) provides an alternative within-position
objective by aligning relative logit geometry, but its complete
pairwise form incurs $\mathcal{O}(|\mathcal{V}|^2)$ complexity.
We propose Bidirectional Centered Score Matching (BCSM), an
efficient within-position distillation objective that constructs
an $\mathcal{O}(K)$ bidirectional active set, reduces pairwise
comparisons to $\mathcal{O}(K^2)$, and factorizes them into an
$\mathcal{O}(K)$ objective. We further introduce cross-timestep
transition alignment. Experiments demonstrate strong
quality--efficiency trade-offs, including up to $6\times$ NFE
reduction with a 16.9\% relative HumanEval gain and a 15.1\%
relative MMMU gain at single-step inference.
\end{abstract}

\section{Introduction}
Discrete diffusion language models~\citep{arriola2025block,nie2025large,ye2025dream} and diffusion vision-language models~\citep{yu2025dimple,li2025lavida,you2026lladav} generate multiple tokens in parallel through iterative denoising, providing a promising alternative to left-to-right autoregressive decoding. 
However, high-quality generation often requires either a large model or multiple denoising steps, which limits the achievable acceleration under aggressive step reduction. 
Knowledge distillation is therefore commonly used to transfer the generation capability of a stronger or more expensive teacher to a student that can operate with fewer evaluations or lower computational cost~\citep{ko2024distillm,wu2025rethinking,wang2026d2f}.

Most existing distillation methods minimize token-wise
Kullback--Leibler (KL) divergence~\citep{ko2024distillm,wu2025rethinking}
between teacher and student predictions at each sequence position.
For parallel decoders, the distribution of a token block is commonly
represented as a product of position-wise categorical distributions.
Accordingly, token-wise KL exactly aligns the corresponding
product-form block distributions, including within-position relative
token preferences.
However, this output factorization cannot represent dependencies
excluded by its cross-position conditional-independence assumption,
and may assign probability mass to combinations that are plausible
position-wise but inconsistent jointly~\citep{gu2018nonautoregressive,zhou2020understanding}.
This representational limitation is shared by all position-wise
objectives, including ours.
Rather than reconstructing the missing joint distribution, we study
an alternative within-position optimization geometry for
few-step discrete diffusion distillation.
Concrete Score Matching (CSM)~\cite{meng2022concrete} directly aligns
relative log-probability ratios between vocabulary states, whose
global partition function cancels analytically, thereby matching the
teacher's within-position relative logit structure.
However, complete pairwise CSM requires
$\mathcal{O}(|\mathcal{V}|^2)$ comparisons, which is costly for large
vocabularies.
Moreover, practical teacher-only truncation creates an asymmetric
candidate-coverage gap~\citep{anshumann2025sparse,wang2021selective,zhong2026duet},
leaving some student-preferred states outside the optimization support.

To address these within-position optimization, computational, and
candidate-coverage challenges, we propose Bidirectional Centered Score
Matching (BCSM), an efficient distillation framework that aligns the
teacher's within-position relative logit geometry without
complete-vocabulary quadratic pairwise matching.
Unlike direct logit matching, BCSM removes additive logit offsets
through mean-centering, allowing the student to match relative token
preferences rather than absolute logit values.
BCSM consists of three integrated mechanisms.
First, we construct dynamic bidirectional active subspaces using a
Kneedle-based maximum-distance criterion to extract adaptive boundaries
for both teacher and student distributions.
This dual formulation incorporates student-specific candidates while
restricting the active set to $\mathcal{O}(K)$ states and the resulting
pairwise comparisons to $\mathcal{O}(K^2)$, where
$K\ll|\mathcal{V}|$.
Second, we introduce a mean-centered formulation that anchors logits
to an active-set weighted mean.
By leveraging the variance identity, we factorize the
$\mathcal{O}(K^2)$ pairwise log-score objective into an
$\mathcal{O}(K)$ pointwise alignment of centered log-scores.
Matching centered log-scores preserves pairwise logit differences and
is invariant to a shared additive shift.
Third, we align centered log-score changes across two noise levels,
providing cross-noise-level regularization for denoising.
We extensively evaluate BCSM on diffusion vision-language models
($e.g.$, Bard-4B-VL~\cite{chen2026bard}) and diffusion language models
($e.g.$, LLaDA-8B~\cite{nie2025large},
Dream-7B~\cite{ye2025dream}).
Empirical results indicate improved local structural consistency and
strong quality--efficiency trade-offs against competitive distillation
baselines across various step budgets.
For example, BCSM achieves $35.67$ on MMMU at $\mathrm{NFE}=1$,
outperforming the undistilled student at $\mathrm{NFE}=32$ ($32.14$),
and scores $42.1$ on HumanEval at $\mathrm{NFE}=42.9$, compared with
$36.0$ for the $256$-step LLaDA-8B base model; see
Section~\ref{subsec:main_results}.
Qualitative examples further show exact reference recovery where KL
baselines produce fragmented or over-extended strings, while additional
cases reveal remaining limitations in long-sequence recovery
(Section~\ref{subsec:discussion}).

\section{Related Work}
\subsection{Knowledge Distillation for Sequence Models}
Knowledge Distillation transfers capabilities from a teacher to a compact student by aligning their output distributions.
Autoregressive models use chain-rule factorization, whereas discrete diffusion models commonly represent parallel predictions as products of position-wise categorical distributions. 
In both cases, KL decomposes according to the chosen factorization. 
Thus, token-wise KL exactly aligns the corresponding product-form distribution but cannot recover dependencies excluded by the parallel output factorization~\cite{gu2018nonautoregressive,zhou2020understanding}. 
Existing objectives, including Forward/Reverse KL, DistiLLM~\cite{ko2024distillm}, and AKL~\cite{wu2025rethinking}, mainly differ in weighting and optimization geometry. 
Alternatively, Direct Logit Distillation applies Mean Squared Error to unnormalized outputs, but absolute logit equivalence restricts the student's solution space~\cite{ba2014deep,urban2017do,ijcai2021p362,sun2024logit}.

\subsection{Score Matching in Discrete Domains}
Score matching optimizes energy-based distributions without evaluating the intractable partition function~\cite{hyvarinen2005estimation,song2021train}. 
While continuous score matching relies on spatial derivatives~\cite{song2023consistency,wang2023prolificdreamer,yin2024one}, these are ill-defined for discrete vocabularies. 
Concrete Score Matching (CSM) instead characterizes discrete energy landscapes through relative probability ratios between states~\cite{meng2022concrete}. 
Aligning log-relative ratios cancels normalization constants and avoids exponential ratio amplification. 
However, a complete pairwise implementation over the vocabulary incurs $\mathcal{O}(|\mathcal{V}|^2)$ complexity, posing a severe computational bottleneck for large-scale models.
\section{Preliminaries}

\subsection{Token-Wise Distillation under Parallel Factorization}
\label{subsec:kl_to_sm}
In standard sequence modeling, knowledge distillation minimizes the KL divergence between teacher and student distributions.
For autoregressive models, the joint probability follows
$P(Y|X)=\prod_{t=1}^{T}P(y_t|y_{<t},X)$,
so token-level KL decomposes consistently with sequence-level optimization.
Parallel discrete models ($e.g.$, block-wise discrete diffusion~\cite{arriola2025block}) instead use product-form probabilities,
$P_{\mathrm{block}}(Y|X)=\prod_{t\in\mathrm{block}}P(y_t|X)$.
Token-wise KL exactly aligns this factorized block distribution, while cross-position dependencies excluded by the factorization remain unmodeled~\cite{gu2018nonautoregressive}.
We next review score matching as an alternative within-position objective under the same factorization.

\subsection{Continuous vs. Discrete Score Matching}
At a fixed prediction position, the categorical distribution can be framed as an Energy-Based Model
$q_{\theta}(y|X)=\frac{\exp\left(-E_{\theta}(y,X)\right)}{Z_{\theta}(X)}$,
where $E_{\theta}(y,X)$ is an unnormalized energy function and
$Z_{\theta}(X)=\sum_{v\in\mathcal{V}}\exp\left(-E_{\theta}(v,X)\right)$
is the partition function.
Score matching~\cite{hyvarinen2005estimation,song2021train} characterizes such unnormalized distributions through quantities in which the shared partition function is eliminated.

\paragraph{Continuous Score Matching.}
\label{subsec:continuous_data}
In continuous spaces ($\mathbf{x}\in\mathbb{R}^d$), taking the spatial derivative (Stein score) of the log-density
$q_{\theta}(\mathbf{x})\propto\exp(-E_{\theta}(\mathbf{x}))$
eliminates the partition function:
$\mathbf{s}_{\theta}(\mathbf{x})
:=\nabla_{\mathbf{x}}\log q_{\theta}(\mathbf{x})
=-\nabla_{\mathbf{x}}E_{\theta}(\mathbf{x})$.
The continuous score matching objective aligns model and data score fields:
\begin{equation}
\mathcal{L}_{\text{SM}}(\theta;p_{\text{data}})
=
\mathbb{E}_{\mathbf{x}\sim p_{\text{data}}}
\left[
\left\|
\nabla_{\mathbf{x}}\log q_{\theta}(\mathbf{x})
-
\nabla_{\mathbf{x}}\log p_{\text{data}}(\mathbf{x})
\right\|_2^2
\right],
\end{equation}
enabling optimization without evaluating $Z_{\theta}$~\cite{wang2023prolificdreamer,yin2024one,song2023consistency}.

\begin{figure*}[t]
    \centering
    \begin{subfigure}[b]{0.20\textwidth}
        \centering
        \parbox[b][4cm][c]{\linewidth}{
            \centering
            $\vcenter{\hbox{\includegraphics[width=\linewidth]{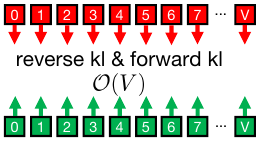}}}$
        }
        \caption{Vanilla KL Divergence.}
        \label{fig:method_vanilla}
    \end{subfigure}
    \hfill
    \begin{subfigure}[b]{0.37\textwidth}
        \centering
        \parbox[b][4cm][c]{\linewidth}{
            \centering
            $\vcenter{\hbox{\includegraphics[width=\linewidth]{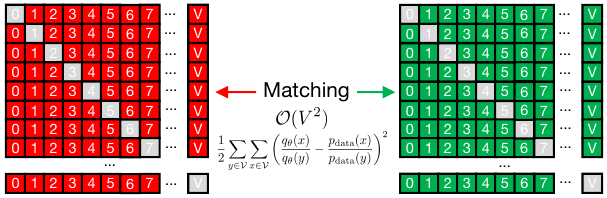}}}$
        }
        \caption{Concrete Score Matching.}
        \label{fig:method_csm}
    \end{subfigure}
    \hfill
    \begin{subfigure}[b]{0.39\textwidth}
        \centering
        \parbox[b][4cm][c]{\linewidth}{
            \centering
            $\vcenter{\hbox{\includegraphics[width=\linewidth]{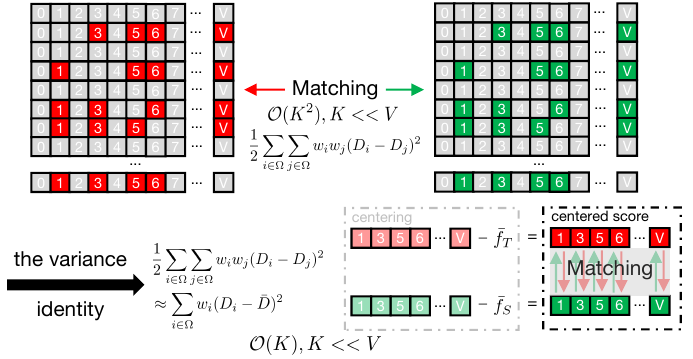}}}$
        }
        \caption{Bidirectional Centered Score Matching.}
        \label{fig:method_bcsm}
    \end{subfigure}

    \caption{\textbf{Conceptual comparison of discrete distillation objectives.}
    (a) Vanilla KL Divergence aligns position-wise categorical probabilities with linear $\mathcal{O}(|\mathcal{V}|)$ complexity.
    (b) Concrete Score Matching aligns pairwise relative probability ratios but incurs $\mathcal{O}(|\mathcal{V}|^2)$ complexity.
    (c) Bidirectional Centered Score Matching restricts evaluation to a dynamic subspace ($K\ll|\mathcal{V}|$) and applies mean-centering factorization, reducing the pairwise matching complexity to $\mathcal{O}(K)$ while preserving relative logit differences within the active set.}
    \label{fig:method_comparison}
    \vspace{-3mm}
\end{figure*}

\paragraph{Discrete Concrete Score Matching.}
\label{subsec:discrete_data}
For discrete states $y\in\mathcal{V}$ with
$q_{\theta}(y)\propto\exp(f_{\theta,y})$,
Concrete Score Matching (CSM) replaces spatial derivatives with the Concrete score
$\mathbf{s}_{\theta}(y):=
\left[q_{\theta}(x)/q_{\theta}(y)\right]_{x\in\mathcal{V}}$,
which characterizes relative probability changes from $y$ to all vocabulary states~\cite{meng2022concrete}.
The partition function cancels analytically:
\begin{equation}
\frac{q_{\theta}(x)}{q_{\theta}(y)}
=
\frac{\exp(f_{\theta,x})/Z_{\theta}}
{\exp(f_{\theta,y})/Z_{\theta}}
=
\exp\left(f_{\theta,x}-f_{\theta,y}\right).
\label{eq:concrete_ratio}
\end{equation}
With uniform pair weights, the complete-vocabulary CSM objective is
\begin{equation}
\mathcal{L}_{\text{CSM}}(\theta;p_{\text{data}})
=
\frac{1}{2}
\sum_{y\in\mathcal{V}}
\sum_{x\in\mathcal{V}}
\left(
\frac{q_{\theta}(x)}{q_{\theta}(y)}
-
\frac{p_{\text{data}}(x)}{p_{\text{data}}(y)}
\right)^2.
\label{eq:csm_objective}
\end{equation}
Matching relative ratios eliminates $Z_{\theta}$, but the complete-vocabulary formulation incurs $\mathcal{O}(|\mathcal{V}|^2)$ pairwise complexity.
\section{Methodology}
Extending score-based distillation to discrete diffusion models presents two bottlenecks.
First, complete-vocabulary ratio matching (Eq.~\ref{eq:concrete_ratio}) requires $\mathcal{O}(|\mathcal{V}|^2)$ pairwise comparisons, which is computationally expensive.
Second, teacher-only truncation creates an asymmetric candidate-coverage gap~\cite{anshumann2025sparse,wang2021selective,zhong2026duet}, leaving student-preferred states outside the optimization support.
To resolve these challenges, we propose Bidirectional Centered Score Matching (see Figure~\ref{fig:method_bcsm}) through three stages:

1) \textbf{dynamic bidirectional active subspaces} that capture both distributions, restricting the active set to $\mathcal{O}(K)$ states and pairwise comparisons to $\mathcal{O}(K^2)$ ($K \ll |\mathcal{V}|$).

2) \textbf{mean-centered discrete score matching} that operates in logit space and reduces the pairwise matching complexity from $\mathcal{O}(K^2)$ to $\mathcal{O}(K)$.

3) \textbf{score transition dynamics extension} that incorporates temporal sequential denoising.

\subsection{Dynamic Bidirectional Active Subspaces}
\label{subsec:dynamic_subspaces}

To reduce pairwise complexity to $\mathcal{O}(K^2)$, we construct an active subspace of size $K \ll |\mathcal{V}|$. 
Standard distillation builds this subset only from teacher predictions $\Omega_T$~\cite{zhong2026duet,anshumann2025sparse}, risking instability when the distributional gap is large. 
We establish a bidirectional active subspace where the teacher preserves target modes and the student incorporates its preferred candidate states.

\begin{figure}[t]
\centering
\includegraphics[width=\linewidth]{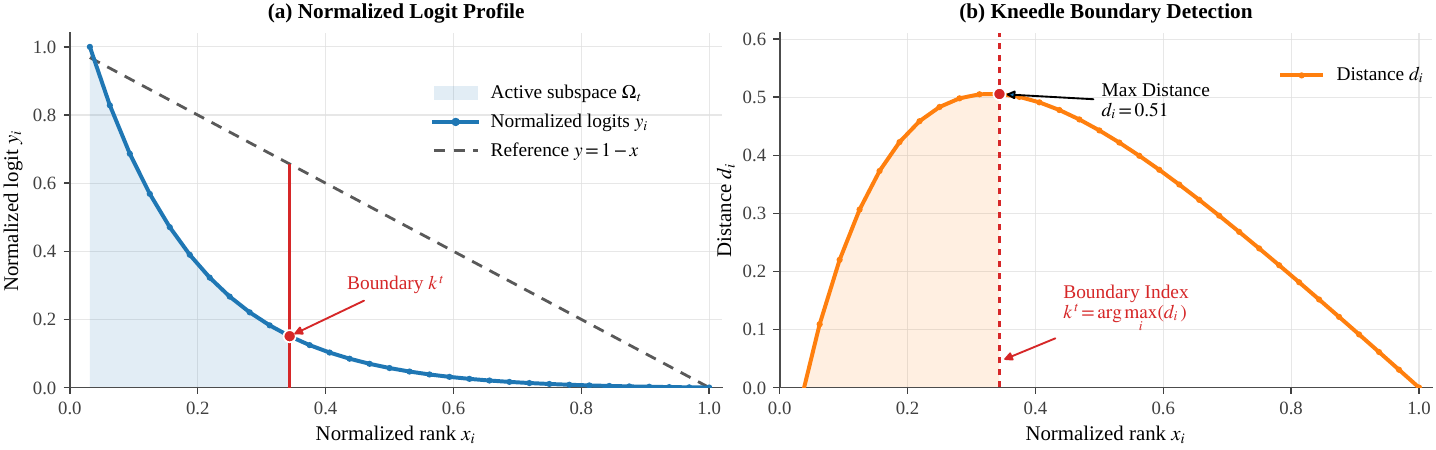}
\caption{\textbf{Kneedle-based extraction of the dynamic active subspace.}
(a) The sorted logits are normalized to form the logit profile
$y_i$ over normalized rank $x_i$; the prefix up to $k^t$ defines the
active subspace $\Omega_t$.
(b) The distance $d_i=(1-x_i)-y_i$ reaches its maximum at
$k^t=\arg\max_i d_i$, which determines the adaptive boundary
(red dotted line).
The same procedure is applied independently to the student logits to
obtain $k^s$ and $\Omega_s$.}
\label{fig:dynamic_subspaces}
\vspace{-3mm}
\end{figure}

Let $f_T,f_S\in\mathbb{R}^{|\mathcal{V}|}$ denote the unnormalized logits. We apply the maximum-distance form of Kneedle~\cite{satopaa2011finding} to locate the transition boundary within a search window $K_{\max}\ll|\mathcal{V}|$. 
Vocabulary sorting is inherited from top-$k$ retrieval and implemented with partial sorting. 
For a sorted vector $\bar{f}$ and $i\in\{1,\ldots,K_{\max}\}$, we define
\begin{equation}
\begin{aligned}
x_i = \frac{i-1}{K_{\max}-1},y_i = \frac{\bar{f}_i-\bar{f}_{K_{\max}}}
{\max\{\bar{f}_1-\bar{f}_{K_{\max}},\epsilon\}},
\end{aligned}
\label{eq:kneedle_normalization}
\end{equation}
where $\epsilon>0$ prevents division by zero for flat logits. The dynamic boundary maximizes the distance to the baseline $y=1-x$, and $k=\arg\max_{i\in\{1,\ldots,K_{\max}\}} \left((1-x_i)-y_i\right)$.
As visualized in Figure~\ref{fig:dynamic_subspaces}, this criterion measures the vertical gap between the normalized logit curve and the diagonal baseline. 
The boundary index ($k^t$ or $k^s$) is identified at the maximum distance, capturing the transition from high-probability modes to the flat tail. Applying this independently to teacher and student distributions yields $k^t$ and $k^s$. 
Their active subspaces are combined as $\Omega=\operatorname{Unique}\!\left(\operatorname{TopK}(f_T,k^t) \cup \operatorname{TopK}(f_S,k^s) \right)$.
The teacher subset preserves principal target modes, while the student subset covers its current preferred states. Consequently, $\Omega$ reduces complete-vocabulary pairwise comparisons from $\mathcal{O}(|\mathcal{V}|^2)$ to $\mathcal{O}(K^2)$, enabling our linear $\mathcal{O}(K)$ formulation.

\subsection{Mean-Centered Discrete Score Matching}
\label{subsec:bcsm}
Based on the relative logit geometry of CSM (Section~\ref{subsec:discrete_data}), we avoid explicitly computing exponential probability ratios by formulating a surrogate pairwise log-ratio objective directly in logit space.
In this space, the discrepancy between states $i,j\in\Omega$ simplifies to $\Delta_{i,j}=D_i-D_j$, where $D_i=f_{S,i}-f_{T,i}$.
We align these relative logit structures by minimizing weighted squared pairwise discrepancies over $\Omega$~\cite{hyvarinen2005estimation}:
\begin{equation}
\mathcal{L}_{\text{Pairwise}}
=
\frac{1}{2W_\Omega}
\sum_{i\in\Omega}\sum_{j\in\Omega}
p_ip_j(D_i-D_j)^2,
\label{eq:pairwise}
\end{equation}
where $p_k=\frac{\exp(f_{S,k})}{\sum_{v\in\mathcal{V}}\exp(f_{S,v})}$, $W_\Omega=\sum_{v\in\Omega}p_v$, and $\alpha_k=p_k/W_\Omega$, requiring $\mathcal{O}(K^2)$ operations.

To reduce the pairwise matching complexity from $\mathcal{O}(K^2)$ to $\mathcal{O}(K)$, we use the variance identity (see Appendix~\ref{appendix:variance_proof} for proof):
\begin{equation}
\frac{1}{2W_\Omega}
\sum_{i,j\in\Omega}p_ip_j(D_i-D_j)^2
=
\sum_{i\in\Omega}p_i(D_i-\bar{D})^2.
\label{eq:variance_identity}
\end{equation}
Here, $\bar{D}=\sum_{k\in\Omega}\alpha_kD_k$.
The baseline shift $\bar{D}=\bar{f}_S-\bar{f}_T$ is defined by the active-set weighted means $\bar{f}_S=\sum_{k\in\Omega}\alpha_kf_{S,k}$ and $\bar{f}_T=\sum_{k\in\Omega}\alpha_kf_{T,k}$.
Substituting $\bar{D}$ into Eq.~\eqref{eq:variance_identity} yields:
\begin{equation}
\sum_{i\in\Omega}p_i
\left(D_i-(\bar{f}_S-\bar{f}_T)\right)^2
=
\sum_{i\in\Omega}p_i
\left(\tilde{f}_{S,i}-\tilde{f}_{T,i}\right)^2,
\end{equation}
where $\tilde{f}_{M,i}=f_{M,i}-\bar{f}_M$ are centered log-scores for $M\in\{S,T\}$, satisfying $\tilde{f}_{M,i}-\tilde{f}_{M,j}=\log q_M(i)-\log q_M(j)$.
This derivation shows that our $\mathcal{O}(K)$ formulation exactly factorizes the proposed $\mathcal{O}(K^2)$ pairwise log-ratio objective.
Operating directly on logits rather than exponential probability ratios avoids exponential amplification of discrepancies while preserving relative differences among active states.
Additionally, centering preserves pairwise differences ($\tilde{f}_{S,i}-\tilde{f}_{S,j}=f_{S,i}-f_{S,j}$) and provides logit-shift invariance~\cite{sun2024logit,ijcai2021p362}, yielding a more flexible solution space than absolute logit matching~\cite{ba2014deep,urban2017do}.

With the factorized discrepancy, we formulate BCSM as a gradient-injected surrogate loss.
The corrective direction for the student logits is defined as the weighted discrepancy $2p_i(\tilde{f}_{S,i}-\tilde{f}_{T,i})$.
For automatic differentiation, the objective is implemented as the inner product between the student logits and the detached corrective direction:
\begin{equation}
\mathcal{L}_{\mathrm{BCSM}}
=
\sum_{i\in\Omega}f_{S,i}\cdot
\operatorname{sg}\!\left(
2p_i\left(\tilde{f}_{S,i}-\tilde{f}_{T,i}\right)
\right),
\label{eq:bcsm_objective}
\end{equation}
where $\operatorname{sg}(\cdot)$ denotes the stop-gradient operator ($e.g.$, \texttt{.detach()} in PyTorch).
The training procedure is summarized in Algorithm~\ref{alg:bcsm} of the Appendix.
While this static objective aligns relative logits at a single timestep, diffusion models operate over sequential timesteps.
The following section extends this foundation to capture temporal score transition dynamics.

\subsection{Formulating BCSM over Discrete Score Transition}
\label{subsec:traj_bcsm}
While Eq.~\eqref{eq:bcsm_objective} aligns relative logits at a single noise level, diffusion models evolve across multiple noise levels.
To capture this evolution, we evaluate teacher and student on the same paired states $x_{t_1}$ and $x_{t_2}$ sampled from one forward diffusion trajectory, where $t_2>t_1$.
The \textit{score transition} denotes the centered-score change between these paired states.
Anchoring the subspace extraction on the lower-noise state $t_1$ via the bidirectional strategy in Section~\ref{subsec:dynamic_subspaces}, we form:
\begin{equation}
\Omega_{\mathrm{trans}}
=
\operatorname{Unique}\!\left(
\operatorname{TopK}(f_{T,t_1},k^t_{t_1})
\cup
\operatorname{TopK}(f_{S,t_1},k^s_{t_1})
\right).
\end{equation}
To maintain stability, we anchor the weights to $t_1$, defining
$p_i=\operatorname{Softmax}_{\mathcal{V}}(f_{S,t_1})_i$ and
$\alpha_i=p_i/\sum_{v\in\Omega_{\mathrm{trans}}}p_v$.
Within $\Omega_{\mathrm{trans}}$, we calculate the centered scores
$\tilde{f}_{S,t_1}$, $\tilde{f}_{S,t_2}$, $\tilde{f}_{T,t_1}$, and
$\tilde{f}_{T,t_2}$ using $\alpha_i$.
The transition vectors are the centered-score differences between the two noise levels:
\begin{equation}
\Delta^{\mathrm{trans}}_S
=
\tilde{f}_{S,t_1}-\tilde{f}_{S,t_2},
\quad
\Delta^{\mathrm{trans}}_T
=
\tilde{f}_{T,t_1}-\tilde{f}_{T,t_2}.
\end{equation}
With the transition discrepancy
$e_{\mathrm{trans},i}
=
\Delta^{\mathrm{trans}}_{S,i}
-
\Delta^{\mathrm{trans}}_{T,i}$,
the alignment loss is
$\mathcal{L}_{\mathrm{trans}}
=
\sum_{i\in\Omega_{\mathrm{trans}}}
p_i e_{\mathrm{trans},i}^{\,2}$.
The total objective is
$\mathcal{L}_{\mathrm{Total}}
=
\mathcal{L}_{\mathrm{Static},t_1}
+
\mathcal{L}_{\mathrm{Static},t_2}
+
\lambda_{\mathrm{trans}}\mathcal{L}_{\mathrm{trans}}$,
where $\mathcal{L}_{\mathrm{Static},t_1}$ and
$\mathcal{L}_{\mathrm{Static},t_2}$ are the static BCSM objectives.
To integrate this joint objective into a surrogate loss, we define the static directions
$g_{\mathrm{static},t,i}
=
2p_i(\tilde{f}_{S,t,i}-\tilde{f}_{T,t,i})$
for $t\in\{t_1,t_2\}$ and the transition direction
$g_{\mathrm{trans},i}
=
2\lambda_{\mathrm{trans}}p_i e_{\mathrm{trans},i}$.
The transition term applies opposite corrections to $t_1$ and $t_2$.
We inject these directions into the global logits via a stop-gradient surrogate loss:
\begin{equation}
\begin{split}
\mathcal{L}_{\mathrm{Dynamic}}
=
\sum_{i\in\Omega_{\mathrm{trans}}}\big[
&f_{S,t_1,i}\cdot\operatorname{sg}
\big(g_{\mathrm{static},t_1,i}+g_{\mathrm{trans},i}\big)\\[-0.5ex]
+&f_{S,t_2,i}\cdot\operatorname{sg}
\big(g_{\mathrm{static},t_2,i}-g_{\mathrm{trans},i}\big)
\big],
\end{split}
\label{eq:dynamic_bcsm_objective}
\end{equation}
The forward pass and gradient injection process is summarized in Algorithm~\ref{alg:dynamic_bcsm} of the Appendix.
\section{Experiments}
\subsection{Experimental Setup}
\paragraph{Implementation Details.}
To assess applicability across modalities,
we evaluate BCSM on both discrete language and vision-language architectures.
For discrete language modeling,
we distill LLaDA-8B-Instruct~\cite{nie2025large} and Dream-7B-Base~\cite{ye2025dream}.
Following the D2F protocol~\cite{wang2026d2f},
these language models are trained on Qwen2.5-7B~\cite{qwen2.5} responses to the Bespoke-Stratos-17k dataset~\cite{bespoke_stratos},
truncated to a maximum sequence length of $600$ tokens.
For vision-language distillation,
we train a student Bard-4B-VL~\cite{chen2026bard} (block size of $32$) from an 8B teacher (block size of $4$) using $170\text{K}$ official image-text pairs.
All models are trained on four NVIDIA H800 (80GB) GPUs,
requiring approximately $23$, $16$, and $9$ hours for Bard-4B-VL, LLaDA-8B-Instruct, and Dream-7B-Base, respectively.
\begin{table}[t]
    \centering
    \scriptsize 
    \renewcommand{\arraystretch}{0.7} 
    \setlength{\tabcolsep}{2.0mm} 
    \caption{\textbf{Performance comparison on diffusion language models.}
    Results for LLaDA-8B-Instruct and Dream-7B-Base on math and coding benchmarks.
    Undistilled base models (NFE=256.0) serve as references. Per CDLM~\cite{kim2025cdlm}, fractional NFEs denote average evaluations per block (BS=256).}
    \label{tab:distillation_results_nfe_dllm}

    \begin{tabular}{ll cccc}
        \toprule
        \multirow{2}{*}{\textbf{Benchmark}} & \multirow{2}{*}{\textbf{Method}} & \multicolumn{2}{c}{\textbf{LLaDA-8B}} & \multicolumn{2}{c}{\textbf{Dream-7B}} \\ 
        \cmidrule(lr){3-4} \cmidrule(lr){5-6} 
        & & \textbf{NFE} $\downarrow$ & \textbf{Score} $\uparrow$ & \textbf{NFE} $\downarrow$ & \textbf{Score} $\uparrow$ \\ 
        \midrule

        \multirow{8}{*}{\begin{tabular}[c]{@{}c@{}}\textbf{GSM8K} \\ (4-shot)\end{tabular}} 
        & Base Model & 256.0 & 77.4 & 256.0 & 75.0 \\
        \cmidrule(lr){2-6}
        & Fast-dLLM
          & 107.5 & \textbf{78.9} & 136.4 & 75.0 \\
        & D2F
          & 87.7 & 77.3 & 128.3 & 77.7 \\
        & CDLM
          & 89.7 & 73.9 & 125.2 & \textbf{78.5} \\
        \cmidrule(lr){2-6}
        & Ours
          & \textbf{86.9} & 77.6 & 126.7 & 77.8 \\
        & Ours$_{\text{Dyn}}$
          & 90.2 & 78.1 & \textbf{78.1} & 78.3 \\
        \midrule
        
        \multirow{8}{*}{\begin{tabular}[c]{@{}c@{}}\textbf{HumanEval} \\ (0-shot)\end{tabular}} 
        & Base Model & 256.0 & 36.0 & 256.0 & 54.3 \\
        \cmidrule(lr){2-6}
        & Fast-dLLM
          & 100.3 & 35.4 & \textbf{72.8} & 53.0 \\
        & D2F
          & 52.8 & 40.2 & 112.3 & 54.3 \\
        & CDLM
          & \textbf{42.3} & 40.2 & 124.3 & 53.7 \\
        \cmidrule(lr){2-6}
        & Ours
          & 42.9 & \textbf{42.1} & 119.6 & 57.3 \\
        & Ours$_{\text{Dyn}}$
          & 51.4 & 40.2 & 110.8 & \textbf{57.9} \\
        \midrule
        
        \multirow{8}{*}{\begin{tabular}[c]{@{}c@{}}\textbf{MATH} \\ (4-shot)\end{tabular}} 
        & Base Model & 256.0 & 23.7 & 256.0 & 35.8 \\
        \cmidrule(lr){2-6}
       & Fast-dLLM
          & 96.9 & 22.4 & 139.8 & \textbf{37.6} \\
        & D2F
          & 142.5 & 29.1 & 133.1 & 35.1 \\
        & CDLM
          & 75.3 & 28.3 & 180.1 & 32.6 \\
        \cmidrule(lr){2-6}
        & Ours
          & 88.8 & \textbf{30.1} & 114.0 & 35.3 \\
        & Ours$_{\text{Dyn}}$
          & \textbf{71.1} & 28.9 & \textbf{113.9} & 35.5 \\
        \midrule
        
        \multirow{8}{*}{\begin{tabular}[c]{@{}c@{}}\textbf{MBPP} \\ (0-shot)\end{tabular}} 
        & Base Model & 256.0 & 39.0 & 256.0 & 56.2 \\
        \cmidrule(lr){2-6}
        & Fast-dLLM
          & 59.1 & 36.4 & 94.6 & 51.0 \\
        & D2F
          & 34.0 & 38.2 & 114.9 & \textbf{54.4} \\
        & CDLM
          & 58.0 & 38.4 & 119.2 & 50.6 \\
        \cmidrule(lr){2-6}
        & Ours
          & 32.6 & 38.6 & 109.6 & 54.0 \\
        & Ours$_{\text{Dyn}}$
          & \textbf{30.7} & \textbf{40.0} & \textbf{74.5} & 53.8 \\
        \bottomrule
    \end{tabular}
    \vspace{-6mm} 
\end{table}
\begin{table}[t]
    \centering
    \scriptsize 
    \renewcommand{\arraystretch}{0.8}
    \setlength{\tabcolsep}{2.5pt} 
    \caption{\textbf{Performance comparison on Bard-4B-VL under varying NFEs.}
    Teacher (BS=4, NFE=4) and undistilled student (BS=32, NFE=32) serve as references; others use BS=32.
    ``OOM'' denotes out of memory.}
    \label{tab:distillation_results_nfe_bardvl}
    
    \begin{tabular}{l cccccc}
        \toprule
        \multirow{2}{*}{\textbf{Method}} & \multicolumn{6}{c}{\textbf{Bard-VL}} \\
        \cmidrule(lr){2-7}
        & \textbf{MMMU} & \textbf{RWQA} & \textbf{AI2D} & \textbf{Math} & \textbf{MMVU} & \textbf{Avg.} \\
        \midrule
        Teacher (NFE=4) & 35.26 & 72.68 & 82.32 & 67.33 & 43.90 & 60.30 \\
        Student (NFE=32) & 32.14 & 68.10 & 79.79 & 61.93 & 37.50 & 55.89 \\
        
        \midrule
        \multicolumn{7}{c}{\textbf{NFE = 1}} \\
        \midrule
        Student & 30.98 & 68.24 & 77.66 & 40.40 & 35.30 & 50.52 \\
        CSM & \multicolumn{6}{c}{OOM} \\ 
        FKL & 32.49 & 69.28 & 79.60 & 50.63 & 31.90 & 52.78 \\
        RKL & 34.28 & 70.72 & 79.18 & 54.20 & 35.80 & 54.84 \\
        DistiLLM & 32.20 & 69.15 & 79.66 & 52.97 & 35.90 & 53.98 \\
        AKL & 33.82 & 70.20 & 79.47 & 48.90 & 38.40 & 54.16 \\
        Ours & 35.67 & \textbf{71.11} & 80.51 & \textbf{58.07} & \textbf{39.20} & \textbf{56.91} \\
        Ours$_{\text{Dyn}}$ & \textbf{35.95} & 70.85 & \textbf{80.93} & 57.97 & 38.70 & 56.88 \\
        
        \midrule
        \multicolumn{7}{c}{\textbf{NFE = 8}} \\
        \midrule
        Student & 31.04 & 68.24 & 79.89 & 55.10 & 38.60 & 54.57 \\
        CSM & \multicolumn{6}{c}{OOM} \\ 
        FKL & 33.01 & 71.11 & 79.95 & 57.63 & 38.80 & 56.10 \\
        RKL & 34.45 & 70.07 & 79.31 & 58.17 & 39.20 & 56.24 \\
        DistiLLM & 33.18 & 69.41 & 79.79 & 58.57 & \textbf{41.60} & 56.51 \\
        AKL & 34.22 & 69.80 & 79.57 & 59.33 & 40.10 & 56.60 \\
        Ours & 35.43 & 71.90 & 80.60 & 61.67 & 41.50 & 58.22 \\
        Ours$_{\text{Dyn}}$ & \textbf{35.49} & \textbf{72.03} & \textbf{80.76} & \textbf{61.70} & 41.40 & \textbf{58.28} \\
        
        \midrule
        \multicolumn{7}{c}{\textbf{NFE = 16}} \\
        \midrule
        Student & 31.45 & 68.76 & 79.70 & 59.97 & 36.90 & 55.36 \\
        CSM & \multicolumn{6}{c}{OOM} \\ 
        FKL & 32.72 & 70.85 & 80.12 & 59.17 & 39.30 & 56.43 \\
        RKL & 34.33 & 70.07 & 79.21 & 59.17 & 39.70 & 56.50 \\
        DistiLLM & 33.18 & 69.54 & 79.99 & 59.53 & 40.80 & 56.61 \\
        AKL & 33.87 & 69.67 & 79.47 & 61.37 & 39.70 & 56.82 \\
        Ours & \textbf{35.61} & 71.50 & 80.51 & \textbf{61.87} & 41.10 & 58.12 \\
        Ours$_{\text{Dyn}}$ & 35.49 & \textbf{72.30} & \textbf{80.80} & 61.73 & \textbf{41.30} &  \textbf{58.32} \\
        \bottomrule
    \end{tabular}
    \vspace{-6mm}
\end{table}
\paragraph{Baselines.}
To evaluate the effectiveness of score matching in mitigating locally inconsistent generations relative to standard divergence minimization, we benchmark our static objective (Ours) and its transition-aware extension (Ours$_{\text{Dyn}}$) against established distillation paradigms.
For language models, baselines comprise Fast-dLLM~\cite{wu2026fastdllm}, D2F~\cite{wang2026d2f}, and CDLM~\cite{kim2025cdlm}.
For vision-language models, comparisons include Forward KL (FKL), Reverse KL (RKL), DistiLLM~\cite{ko2024distillm}, and Adaptive KL (AKL)~\cite{wu2025rethinking}.
\paragraph{Benchmarks and Metrics.}
Textual capabilities are assessed on GSM8K/GSM8K-CoT~\cite{cobbe2021gsm8k}, MATH~\cite{hendrycks2021math}, HumanEval~\cite{chen2021humaneval}, and MBPP~\cite{austin2021mbpp}.
Coding tasks are evaluated in a zero-shot setting, whereas mathematical benchmarks utilize standard few-shot configurations~\cite{nie2025large,ye2025dream,wang2026d2f}.
Multimodal capabilities are evaluated across six distinct dimensions: multi-image comprehension (MMMU-Pro~\cite{yue-etal-2025-mmmu}, denoted as MMMU), spatial reasoning (RealWorldQA~\cite{xai2024grok15v}, RWQA), video processing (MMVU \textit{val}~\cite{zhao2025mmvu}), diagram interpretation (AI2D~\cite{kembhavi2016diagram}), document understanding (OCRBench~\cite{Liu_2024}, OCR), and mathematical vision reasoning (MathVista \textit{testmini}~\cite{lu2024mathvista}, Math), with OCRBench used for ablation studies.
Finally, generation efficiency is measured by the Number of Function Evaluations (NFE) on a unified dual-H800 GPU setup under single-batch data parallelism.

\begin{table*}[t]
    \centering
    \scriptsize 
    \renewcommand{\arraystretch}{0.6} 
    \setlength{\tabcolsep}{1.2pt} 
    
    \caption{\textbf{Comprehensive ablation studies of BCSM on the OCR benchmark.}}
    \label{tab:comprehensive_ablation}
    
    \begin{minipage}[t]{0.48\textwidth}
        \centering
        \begin{subtable}{\linewidth}
            \centering
            \renewcommand{\arraystretch}{0.8} 
            \caption{Necessity and design choices of the mean-centering mechanism.}
            \label{tab:ablation_centering}
            \begin{tabular*}{\linewidth}{@{\extracolsep{\fill}}l c cccc@{}} 
                \toprule
                & \textbf{Baseline} & \multicolumn{4}{c}{\textit{Core \& Format Alternatives} ($\text{NFE} = 16$)} \\
                \cmidrule(lr){3-6}
                \textbf{Benchmark} & \begin{tabular}[c]{@{}c@{}}Student\\($\text{NFE} = 32$)\end{tabular} & \begin{tabular}[c]{@{}c@{}}No Centering\\(Abs. Logits)\end{tabular} & \begin{tabular}[c]{@{}c@{}}Local\\Softmax KD\end{tabular} & \begin{tabular}[c]{@{}c@{}}Global\\Softmax KD\end{tabular} & \begin{tabular}[c]{@{}c@{}}Ours\end{tabular} \\ 
                \midrule
                \textbf{OCR} & 70.20 & 59.60 & 65.00 & 71.10 & 71.30 \\
                \bottomrule
            \end{tabular*}
        \end{subtable}


        \begin{subtable}{\linewidth}
            \centering
            \renewcommand{\arraystretch}{0.8} 
            \begin{tabular*}{\linewidth}{@{\extracolsep{\fill}}l c ccc@{}} 
                \toprule
                & \textbf{Baseline} & \multicolumn{3}{c}{\textit{Centering Strategies} ($\text{NFE} = 16$)} \\
                \cmidrule(lr){3-5}
                \textbf{Benchmark} & \begin{tabular}[c]{@{}c@{}}Student\\($\text{NFE} = 32$)\end{tabular} & \begin{tabular}[c]{@{}c@{}}Uniform\\Weight\end{tabular} & \begin{tabular}[c]{@{}c@{}}Local\\Weighted\end{tabular} & \begin{tabular}[c]{@{}c@{}}Ours\end{tabular} \\ 
                \midrule
                \textbf{OCR} & 70.20 & 51.90 & 55.10 & 71.30 \\
                \bottomrule
            \end{tabular*}
        \end{subtable}
    \end{minipage}%
    \hfill
    \begin{minipage}[t]{0.48\textwidth}
        \centering
        \begin{subtable}{\linewidth}
            \centering
            \renewcommand{\arraystretch}{0.7} 
            \caption{Components of the Dynamic Bidirectional Active Subspaces.}
            \label{tab:subspace_ablation}
            \begin{tabular*}{\linewidth}{@{\extracolsep{\fill}}l ccccc c@{}} 
                \toprule
                \multirow{3}{*}{\textbf{Benchmark}} & \multicolumn{6}{c}{\textit{Teacher-Only Subspace} ($\Omega_T$, $\text{NFE} = 16$)} \\
                \cmidrule(lr){2-7}
                & \multicolumn{5}{c}{Fixed} & \multirow{2}{*}{Dynamic} \\
                \cmidrule(lr){2-6}
                & Top-64 & Top-128 & Top-256 & Top-512 & Top-1024 & \\ 
                \midrule
                \textbf{OCR} & 62.80 & 63.40 & 64.20 & 63.10 & 62.50 & 68.60 \\
                \bottomrule
            \end{tabular*}
        \end{subtable}


        \begin{subtable}{\linewidth}
            \centering
            \renewcommand{\arraystretch}{0.7} 
            \begin{tabular*}{\linewidth}{@{\extracolsep{\fill}}l ccccc c@{}} 
                \toprule
                \multirow{3}{*}{\textbf{Benchmark}} & \multicolumn{6}{c}{\textit{Bidirectional Subspace} ($\Omega_T \cup \Omega_S$, $\text{NFE} = 16$)} \\
                \cmidrule(lr){2-7}
                & \multicolumn{5}{c}{Fixed} & \multirow{2}{*}{Dynamic} \\
                \cmidrule(lr){2-6}
                & Top-64 & Top-128 & Top-256 & Top-512 & Top-1024 & \\ 
                \midrule
                \textbf{OCR} & 64.50 & 64.00 & 66.60 & 65.10 & 65.40 & 71.30 \\
                \bottomrule
            \end{tabular*}
        \end{subtable}
    \end{minipage}

    \bigskip

    \begin{minipage}[t]{0.32\textwidth}
        \centering
        \begin{subtable}{\linewidth}
            \centering
            \caption{Impact of the search window size $K_{\max}$.}
            \label{tab:kmax_ablation}
            \begin{tabular*}{\linewidth}{@{\extracolsep{\fill}}l cccccc@{}} 
                \toprule
                \multirow{2}{*}{\textbf{Bench}} & \multicolumn{6}{c}{\textit{$K_{\max}$ }($\text{NFE}=16$)} \\
                \cmidrule(lr){2-7}
                & 64 & 128 & 256 & 512 & 768 & 1024~(ours) \\ 
                \midrule
                \textbf{OCR} & 65.4 & 67.3 & 66.6 & 69.6 & 70.9 & 71.3\\
                \bottomrule
            \end{tabular*}
        \end{subtable}
    \end{minipage}%
    \hfill
    \begin{minipage}[t]{0.3\textwidth}
        \centering
        \begin{subtable}{\linewidth}
            \centering
            \caption{Transition weight sensitivity $\lambda_{\text{trans}}$.}
            \label{tab:lambda_traj_ablation}
            \begin{tabular*}{\linewidth}{@{\extracolsep{\fill}}l ccccc@{}} 
                \toprule
                \multirow{2}{*}{\textbf{Bench}}
                & \multicolumn{5}{c}{%
                    \makebox[0pt][c]{%
                        \textit{Transition Weight}
                        ($\lambda_{\text{trans}}$, $\text{NFE}=16$)%
                    }%
                } \\
                \cmidrule(lr){2-6}
                & 0.0 & 0.05 & 0.1 & 0.15 & 0.2 \\ 
                \midrule
                \textbf{OCR} & 71.3 & 71.5 & 71.9 & 70.5 & 69.0  \\
                \bottomrule
            \end{tabular*}
        \end{subtable}
    \end{minipage}%
    \hfill
    \begin{minipage}[t]{0.32\textwidth}
        \centering
        \renewcommand{\arraystretch}{0.5}
        \begin{subtable}{\linewidth}
            \centering
            \caption{Ablation on transition alignment.}
            \label{tab:traj_loss_ablation}
            \begin{tabular*}{\linewidth}{@{\extracolsep{\fill}}l cc cc cc@{}}
                \toprule
                \multirow{2}{*}{\textbf{Bench}}
                & \multicolumn{2}{c}{$\text{NFE}=1$}
                & \multicolumn{2}{c}{$\text{NFE}=8$}
                & \multicolumn{2}{c}{$\text{NFE}=16$} \\
                \cmidrule(lr){2-3}\cmidrule(lr){4-5}\cmidrule(lr){6-7}
                & Ours & +Trans. & Ours & +Trans. & Ours & +Trans. \\
                \midrule
                \textbf{OCR} & 40.9 & 40.9 & 65.9 & 66.8 & 71.3 & 71.9 \\
                \bottomrule
            \end{tabular*}
        \end{subtable}
    \end{minipage}  
\end{table*}
\begin{figure}[t]
    \centering
     \includegraphics[width=0.9\linewidth]{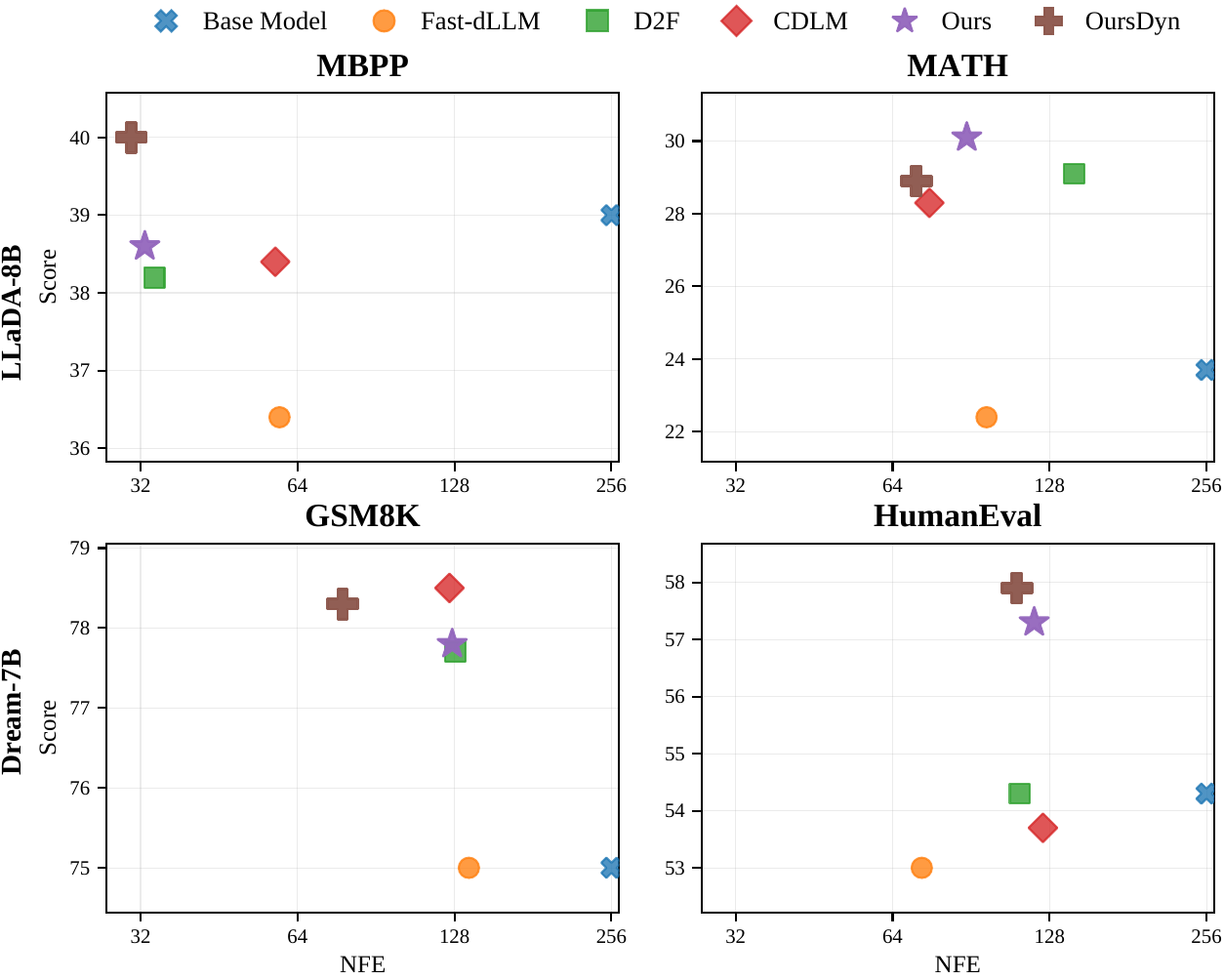}
    \caption{\textbf{Efficiency and performance comparison on diffusion language models.} The top row reports LLaDA-8B results on MBPP and MATH, while the bottom row reports Dream-7B results on GSM8K and HumanEval. Each marker denotes a distillation method, with the number of function evaluations (NFE) on the x-axis and task performance on the y-axis. Lower NFE and higher scores are preferred. Across the four benchmarks, BCSM lies on the efficiency--performance Pareto frontier, attaining higher scores at similar NFE and comparable scores with substantially fewer evaluations.}
    \label{fig:efficiency_tradeoff}
\end{figure}
\subsection{Main Results}
\label{subsec:main_results}
\paragraph{Evaluation on Diffusion Language Models.}
Following the D2F protocol,
we evaluate LLaDA-8B and Dream-7B on coding and mathematical reasoning using fractional NFEs.
As detailed in Table~\ref{tab:distillation_results_nfe_dllm},
BCSM achieves strong quality--efficiency trade-offs relative to undistilled full-step models and competitive distillation baselines.
On HumanEval, our static formulation (Ours) achieves $42.1$ at $\text{NFE}=42.9$ on LLaDA-8B,
outperforming both the base model ($36.0$) and CDLM ($40.2$).
Similarly, the dynamic variant (Ours$_{\text{Dyn}}$) reaches $57.9$ at $\text{NFE}=110.8$ on Dream-7B,
exceeding the full-step baseline.
These quality--efficiency gains also extend to the more challenging MATH benchmark, where Ours attains $30.1$ at $\text{NFE}=88.8$ on LLaDA-8B,
outperforming the base model ($23.7$) and D2F ($29.1$ at $\text{NFE}=142.5$).
Figure~\ref{fig:efficiency_tradeoff} further visualizes these quality--efficiency trade-offs across the evaluated benchmarks.

\paragraph{Evaluation on Vision-Language Models.}
To preserve parallel decoding speedups while mitigating locally inconsistent token combinations,
we employ cross-block distillation (teacher $\text{BS}=4 \rightarrow$ student $\text{BS}=32$) on Bard-4B-VL.
Table~\ref{tab:distillation_results_nfe_bardvl} shows that our framework achieves the best average score across all tested step budgets.
At $\text{NFE}=1$,
both BCSM variants surpass the undistilled 32-step student reference on the general benchmark MMMU.
On Math, Ours$_{\text{Dyn}}$ reaches $61.73$ at $\text{NFE}=16$,
nearly matching the 32-step student ($61.93$) while outperforming RKL and AKL.
At $\text{NFE}=8$, both variants exceed the undistilled 32-step student on RWQA and MMVU.
The complete-vocabulary CSM baseline runs out of memory across all tested settings,
illustrating the practical cost of its $\mathcal{O}(|\mathcal{V}|^2)$ pairwise formulation.

\subsection{Ablation Studies}
We ablate BCSM's core components on Bard-4B-VL using the OCR task, a challenging benchmark with multi-token outputs, by investigating:
1) the necessity of probability-weighted mean-centering over absolute-logit or post-softmax matching;
2) the bidirectional active subspace for addressing asymmetric candidate coverage;
3) dynamic boundary robustness to $K_{\max}$;
4) cross-timestep transition alignment across $\text{NFE}$ regimes; and
5) sensitivity to the transition weight ($\lambda_{\text{trans}}$).
Unless otherwise specified, variants are evaluated at $\text{NFE}=16$ against the undistilled $\text{NFE}=32$ student reference.

\paragraph{Necessity of Mean-Centering.}
We design two groups of ablations to isolate whether the gains arise from the matching representation or the weighting strategy (Table~\ref{tab:ablation_centering}).
First, we compare mean-centering with three matching formats on the same active subspace $\Omega$.
\textit{No Centering (Abs. Logits)} applies Direct Logit Distillation~\cite{ba2014deep,urban2017do,ijcai2021p362}, matching $f_{S,i}$ and $f_{T,i}$ directly to test the necessity of removing additive logit shifts.
\textit{Local Softmax KD} applies Softmax separately to teacher and student logits within $\Omega$ and matches the resulting local distributions, testing whether local probability matching alone suffices.
\textit{Global Softmax KD} applies Softmax over the full vocabulary $\mathcal{V}$ and matches the probabilities indexed by $\Omega$, testing the effect of preserving full-vocabulary normalization.
As shown in Table~\ref{tab:ablation_centering} (top), \textit{No Centering} yields $59.60$ on OCR.
While \textit{Global Softmax KD} reaches $71.10$, our mean-centered method achieves $71.30$, exceeding the undistilled student reference ($70.20$).
Second, we vary the weights used for centering and loss computation while retaining centered-logit matching.
\textit{Uniform Weight} uses $1/|\Omega|$ for both operations, testing whether token probabilities are necessary, and yields $51.90$.
\textit{Local Weighted} applies Softmax to student logits within $\Omega$ and uses the resulting probabilities for both centering and loss weighting, testing whether locally normalized confidence is sufficient, and yields $55.10$.
Our method computes full-vocabulary probabilities $p_i=\operatorname{Softmax}_{\mathcal{V}}(f_S)_i$, uses $\alpha_i=p_i/\sum_{v\in\Omega}p_v$ for centering, and retains $p_i$ for loss weighting, preserving the active set's global probability mass and achieving $71.30$.
These results support both mean-centering and probability-mass-aware weighting.

\begin{figure}[t]
    \centering
    \includegraphics[width=0.85\columnwidth]{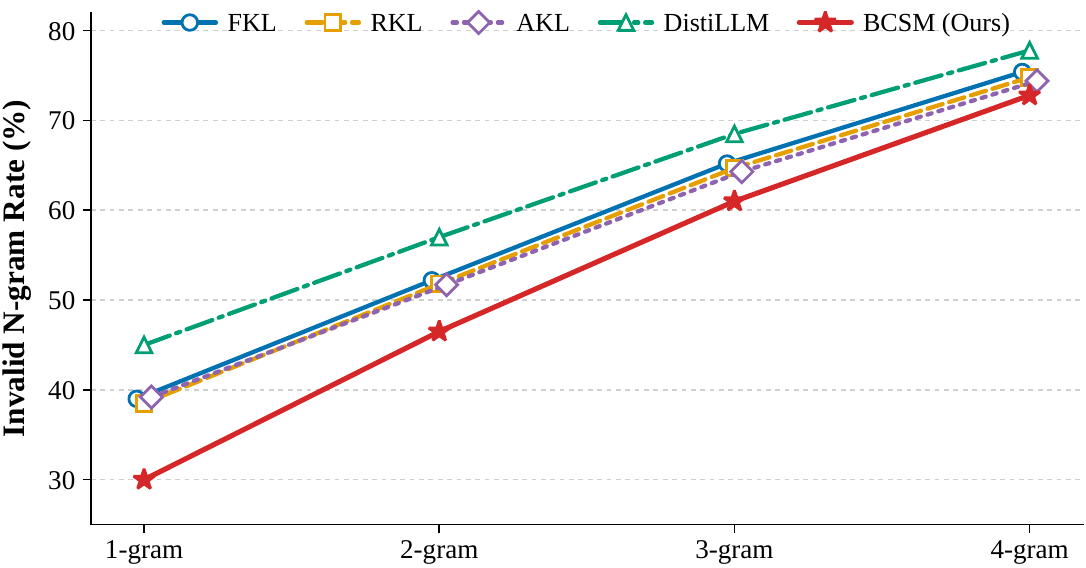}
    \caption{\textbf{Comparison of non-reference n-gram rates (lower is better).}
    BCSM yields lower reference mismatch across all n-gram orders than the baselines, providing empirical evidence of improved local consistency.}
    \label{fig:leakage}
\end{figure}
\paragraph{Impact of Bidirectional Subspace.}
We evaluate the effect of incorporating student-side candidates and dynamic boundary selection in the active subspace construction.
First, we compare the bidirectional subspace with teacher-only selection under both fixed and dynamic settings.
The teacher-only strategy restricts optimization to states selected from the teacher distribution, while the bidirectional strategy additionally includes student candidate states to enlarge the optimization support.
Under the dynamic setting, the bidirectional subspace achieves an OCR score of $71.30$, outperforming the teacher-only counterpart ($68.60$).
This demonstrates that relying solely on teacher-selected states cannot fully cover student-side candidate states, leaving some student errors outside the optimization support.
By incorporating student candidates, the bidirectional strategy provides more complete supervision over the student's current prediction space.
Second, we compare dynamic boundary selection against fixed Top-$K$ configurations.
Within the bidirectional subspace, fixed Top-$K$ variants achieve scores between $64.00$ and $66.60$, while the dynamic strategy reaches $71.30$.
This indicates that adaptive boundary selection better identifies the transition between high-probability modes and low-information tail states, avoiding both insufficient coverage and unnecessary expansion of the active subspace.

\paragraph{Robustness of Dynamic Boundary Selection.}
To assess the robustness of dynamic boundary selection, we vary the preliminary search window $K_{\max}$ in Table~\ref{tab:kmax_ablation}.
As $K_{\max}$ increases from $64$ to $768$, the OCR score generally improves from $65.40$ to $70.90$, and reaches $71.30$ at the default setting of $1024$.
The small gain from $768$ to $1024$ indicates that performance has largely saturated once the search window is sufficiently large.
This suggests that a sufficient initial window enables dynamic boundary selection to cover relevant candidate states without strong sensitivity near the default setting.
Thus, we use $K_{\max}=1024$ by default.

\paragraph{Effectiveness of Transition Alignment.}
We evaluate transition alignment across different NFE regimes in Table~\ref{tab:traj_loss_ablation}.
At $\text{NFE}=1$, models with and without transition alignment both achieve $40.90$ on OCR.
At $\text{NFE}=8$, transition alignment increases the score from $65.90$ to $66.80$.
At $\text{NFE}=16$, the score increases from $71.30$ to $71.90$.
These results indicate no gain at single-step inference but consistent improvements in multi-step generation.

\paragraph{Transition Weight Sensitivity.}
Finally, we examine the impact of the transition weight $\lambda_{\mathrm{trans}}$ in Eq.~\eqref{eq:dynamic_bcsm_objective} at $\mathrm{NFE}=16$ in Table~\ref{tab:lambda_traj_ablation}.
Setting $\lambda_{\mathrm{trans}}=0$ relies solely on static alignment and yields $71.30$ on OCR.
Increasing $\lambda_{\mathrm{trans}}$ to $0.05$ and $0.1$ improves the score to $71.50$ and $71.90$, respectively.
Further increasing it to $0.15$ and $0.2$ reduces the score to $70.50$ and $69.00$.
These results suggest that moderate transition weighting complements static alignment, whereas excessive weighting can weaken it.
Thus, we set $\lambda_{\mathrm{trans}}=0.1$ by default.

\subsection{Discussion}
\label{subsec:discussion}
\begin{figure}[t]
    \centering
    \includegraphics[width=\columnwidth]{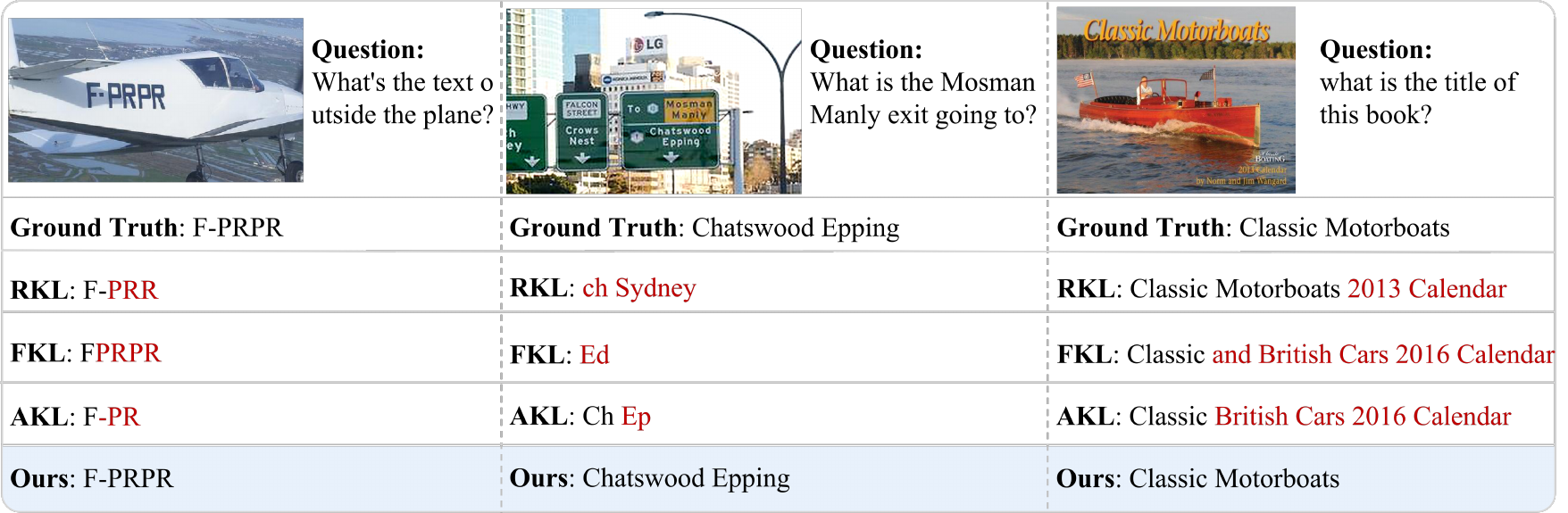}
    \caption{\textbf{Qualitative comparison of generated responses ($\mathrm{NFE}=16$).}
    Red text indicates incorrect or fragmented spans.}
    \label{fig:qualitative_comparison}
\end{figure}

\paragraph{Empirical Analysis of Local Generation Consistency.}
To analyze local generation consistency across position-wise distillation objectives, we evaluate on the MS COCO Caption 2014 validation set (20,252 samples).
Image captioning requires structured text and provides a controlled setting for measuring agreement with reference captions.
We define the \textit{non-reference n-gram rate} as $100(1-p_n)\%$, where $p_n$ is BLEU's clipped n-gram precision ($n\in\{1,\ldots,4\}$).
Because clipping discounts excessive repetitions and n-grams unsupported by references, this metric measures local reference mismatch rather than invalidity directly.
As Figure~\ref{fig:leakage} illustrates, BCSM consistently yields the lowest non-reference n-gram rates across all levels compared with FKL, RKL, DistiLLM, and AKL.
Specifically, BCSM reduces the 1-gram mismatch rate to $30.2\%$, versus $38.6\%$--$45.0\%$ for the baselines.
This trend persists at the 4-gram level, where BCSM achieves $72.7\%$ while all baselines exceed $74.5\%$.
Complementing this corpus-level analysis, Figure~\ref{fig:qualitative_comparison} presents OCR examples where KL baselines produce fragmented or over-extended strings, whereas BCSM exactly recovers the references in all three cases.
Together, these results indicate improved local reference agreement and structural preservation under BCSM, without implying explicit cross-position dependency modeling (see Appendix~\ref{app:probability_leakage} for additional visualizations).

\noindent\paragraph{Failure Cases and Limitations.}
We further examine cases where models fail to recover the exact reference (Figure~\ref{fig:divergent_generations} in the Appendix).
In these cases, FKL and RKL produce fragmented tokens ($e.g.$, ``Ed'') or irrelevant text ($e.g.$, ``TABLE II'').
BCSM often retains more coherent partial structures when the exact reference is missed.
For example, outputs such as ``Chatswood ...'' preserve recognizable local structure.
Nevertheless, BCSM still struggles with exact long-sequence recovery, occasionally producing truncated outputs or minor character deviations (see Appendix~\ref{app:semantic_stability} for additional qualitative comparisons).

\section{Conclusion}
We introduce Bidirectional Centered Score Matching (BCSM), an efficient within-position distillation framework for discrete diffusion models.
By combining dynamic bidirectional active subspaces with mean-centered factorization, BCSM reduces the active-set pairwise matching complexity from $\mathcal{O}(K^2)$ to $\mathcal{O}(K)$ and aligns centered-score transitions across noise levels.
Experiments on language and vision-language models demonstrate strong quality--efficiency trade-offs across inference budgets, together with lower local reference mismatch and improved structural preservation in qualitative examples, while exact long-sequence recovery remains challenging.

\bibliography{custom}

\clearpage
\appendix
\begin{figure*}[t]
    \centering
    \includegraphics[width=0.85\textwidth]{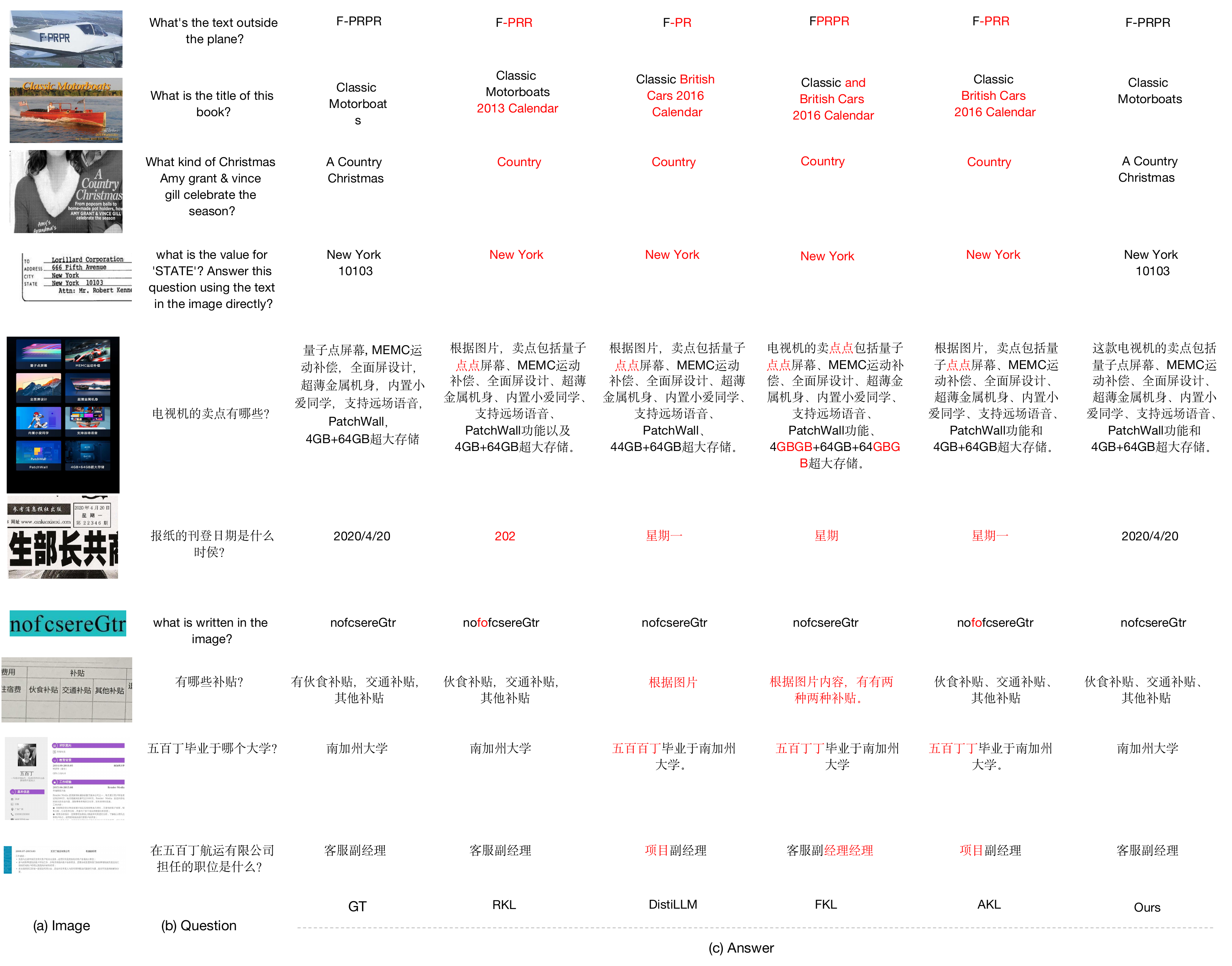}
    \caption{Qualitative comparison of generated responses across distillation methods. In these examples, the KL-based baselines exhibit spelling errors, token repetitions, or malformed compositions, whereas BCSM produces outputs with higher reference agreement. These examples illustrate empirical differences in output quality and do not imply explicit recovery of cross-position dependencies.}
    \label{fig:probability_leakage_vis}
\end{figure*}
 \section{Appendix}
\label{sec:appendix}

\subsection{Additional Training Details}
\label{appendix:training_details}

\paragraph{Training On BARD-VL.}
\paragraph{Training On BARD-VL.}
To adapt our approach to the mask-diffusion version of BARD-VL, we retain its memory-friendly packed sequence construction and standard absorbing-state masked diffusion process~\cite{chen2026bard}, while replacing the original token-wise KL distillation objective with Eq.~\ref{eq:bcsm_objective} or Eq.~\ref{eq:dynamic_bcsm_objective}:

\begin{itemize}
    \item \textit{Memory-Friendly Sequence Construction.} Following BARD, to mitigate memory limits caused by long visual contexts, a packed sequence layout $\tilde{x} = [Q, x_1, x_t]$ is utilized, where $Q$ is the multimodal context, $x_1$ is the clean response, and $x_t$ is the corrupted response. The context $Q$ is stored once, with a causal mask applied to $Q$ and $x_1$. 
    The attention scope for the $i$-th corrupted block $x_t^i$ is defined as $Attn(x_t^i) = Q \cup \{x_1^0, ..., x_1^{i-1}\} \cup x_t^i$, preventing information leakage and enabling block-level parallel computation.
    
    \item \textit{Masked Diffusion Scheduler.} We employ standard absorbing-state masked diffusion, in which completion tokens are corrupted exclusively by the \texttt{[MASK]} token. 
    Given a clean completion $x_1$, a timestep $t\sim\mathcal{U}(0,1)$ is sampled, and each completion token is independently replaced by \texttt{[MASK]} according to the corresponding masking probability. 
    The supervised positions are defined as $\mathcal{M}_{t}=\{i\in\mathcal{C}\mid x_{t,i}=\texttt{[MASK]}\}$.
    
    \item \textit{BCSM Distillation Process.} Building upon this setup, we replace the original KL divergence with the proposed BCSM objective. 
    The teacher is a fixed BARD-VL-4B model with block size $4$, while the student is a BARD-VL-4B model with block size $32$. 
    We directly perform single-stage cross-block distillation without progressive block merging or repeated stage-wise distillation.
    Both models process the same packed sequence $\tilde{x}$ and the same corrupted response $x_t$. 
    Distillation is performed strictly at the masked positions in $\mathcal{M}_t$. 
    The unnormalized teacher and student logits, denoted by $f_T$ and $f_S$, are extracted at these positions. 
    Kneedle determines the dynamic bidirectional active subspace, and the mean-centered variance identity computes the $\mathcal{O}(K)$ BCSM objective used to update the student.
    
\end{itemize}

\begin{algorithm}[t]
    \scriptsize
    \SetAlgoLined
    \SetAlCapSkip{0.5em}
    \setlength\belowcaptionskip{-5pt}
    \caption{Bidirectional Centered Score Matching (BCSM) Distillation}
    \label{alg:bcsm}
    
    \KwIn{Pretrained discrete diffusion teacher $\theta_T$, student model $\theta_S$, training set $\mathcal{D}$, max search window $K_{\max}$, learning rate $\eta$}
    \KwOut{Distilled student model $\theta_S$}

    freeze teacher model $\theta_T$\;
    \While{not converged}{
        \tcp{Step 1: Sample data and compute logits}
        sample clean data $x_0$ from $\mathcal{D}$\;
        sample timestep $t \sim \mathrm{U}(1, T)$ and corrupted state $x_t \sim q(x_t \mid x_0, t)$\;
        compute teacher unnormalized logits $f_T \leftarrow \mathrm{Model}_{\theta_T}(x_t, t)$\;
        compute student unnormalized logits $f_S \leftarrow \mathrm{Model}_{\theta_S}(x_t, t)$\;
        compute global student probability $\mathbf{w} \leftarrow \mathrm{Softmax}(f_S)$\;
        
        \tcp{Step 2: Construct Dynamic Bidirectional Active Subspace}
        compute dynamic boundaries $k^t, k^s \le K_{\max}$ via Kneedle algorithm on $f_T, f_S$\;
        extract active subsets $\Omega_T \leftarrow \mathrm{TopK}(f_T, k^t)$ and $\Omega_S \leftarrow \mathrm{TopK}(f_S, k^s)$\;
        form global active subspace $\Omega \leftarrow \mathrm{Unique}(\Omega_T \cup \Omega_S)$\;

        \tcp{Step 3: Mean-Centered Discrete Score Matching}
        compute truncated weight sum $W_\Omega \leftarrow \sum_{k \in \Omega} w_k$\;
        \For{\(\mathrm{model} \in \{T, S\}\)}{
            compute local expected energy $\bar{f}_{\mathrm{model}} \leftarrow \frac{1}{W_\Omega} \sum_{k \in \Omega} w_k f_{\mathrm{model}, k}$\;
            compute centered scores $\tilde{f}_{\mathrm{model}, i} \leftarrow f_{\mathrm{model}, i} - \bar{f}_{\mathrm{model}}$ for $\forall i \in \Omega$\;
        }
        
        \tcp{Step 4: Compute Loss and Update Student}
        compute objective $\mathcal{L}_{\mathrm{BCSM}} \leftarrow \sum_{i \in \Omega} f_{S,i} \cdot \text{sg}\Big( 2w_i \left( \tilde{f}_{S,i} - \tilde{f}_{T,i} \right) \Big)$\;
        update student model $\theta_S \leftarrow \theta_S - \eta \nabla_{\theta_S} \mathcal{L}_{\mathrm{BCSM}}$\;
    }
    \KwRet{$\theta_S$}
\end{algorithm}

\begin{algorithm}[t]
    \SetAlgoLined
    \SetAlCapSkip{0pt}
    \setlength{\belowcaptionskip}{-2pt}
    \caption{Dynamic BCSM Distillation with Transition Alignment}
    \label{alg:dynamic_bcsm}

    \fontsize{7pt}{7.4pt}\selectfont
    \setlength{\parskip}{0pt}

    \KwIn{Pretrained teacher $\theta_T$, student model $\theta_S$, dataset $\mathcal{D}$, max search window $K_{\max}$, learning rate $\eta$, transition weight $\lambda_{\mathrm{trans}}$}
    \KwOut{Distilled student model $\theta_S$}

    freeze teacher model $\theta_T$\;
    \While{not converged}{
        \tcp{Step 1: Sample two states along the same trajectory}
        sample clean data $x_0\sim\mathcal{D}$\;
        sample timesteps $t_1\sim\mathrm{U}(1,T-1)$ and $t_2>t_1$\;
        sample $x_{t_1}\sim q_{t_1|0}(x_{t_1}\mid x_0)$, then continue the same forward diffusion trajectory to sample $x_{t_2}\sim q_{t_2|t_1}(x_{t_2}\mid x_{t_1})$\;
        compute teacher logits
        $f_{T,t_1},f_{T,t_2}\leftarrow
        \mathrm{Model}_{\theta_T}(x_{t_1},t_1),
        \mathrm{Model}_{\theta_T}(x_{t_2},t_2)$\;
        compute student logits
        $f_{S,t_1},f_{S,t_2}\leftarrow
        \mathrm{Model}_{\theta_S}(x_{t_1},t_1),
        \mathrm{Model}_{\theta_S}(x_{t_2},t_2)$\;

        \tcp{Step 2: Construct the bidirectional active subspace at $t_1$}
        compute dynamic boundaries $k^t_{t_1},k^s_{t_1}\leq K_{\max}$ via the Kneedle-based maximum-distance criterion on $f_{T,t_1}$ and $f_{S,t_1}$\;
        $\Omega_{\mathrm{trans}}\leftarrow
        \operatorname{Unique}\!\left(
        \operatorname{TopK}(f_{T,t_1},k^t_{t_1})
        \cup
        \operatorname{TopK}(f_{S,t_1},k^s_{t_1})
        \right)$\;

        \tcp{Step 3: Compute $t_1$-anchored weights and centered scores}
        compute full-vocabulary probabilities
        $p_i\leftarrow\operatorname{Softmax}_{\mathcal{V}}(f_{S,t_1})_i$\;
        compute
        $W_{\Omega}\leftarrow\sum_{k\in\Omega_{\mathrm{trans}}}p_k$
        and
        $\alpha_i\leftarrow p_i/W_{\Omega}$
        for all $i\in\Omega_{\mathrm{trans}}$\;

        \For{\(\mathrm{model}\in\{T,S\}\) and \(t\in\{t_1,t_2\}\)}{
            compute weighted mean
            $\bar{f}_{\mathrm{model},t}\leftarrow
            \sum_{k\in\Omega_{\mathrm{trans}}}
            \alpha_k f_{\mathrm{model},t,k}$\;
            compute centered scores
            $\tilde{f}_{\mathrm{model},t,i}\leftarrow
            f_{\mathrm{model},t,i}-\bar{f}_{\mathrm{model},t}$
            for all $i\in\Omega_{\mathrm{trans}}$\;
        }

        \tcp{Step 4: Compute static and transition directions}
        \For{\(\forall i\in\Omega_{\mathrm{trans}}\)}{
            $g_{\mathrm{static},t_1,i}\leftarrow
            2p_i\big(
            \tilde{f}_{S,t_1,i}-\tilde{f}_{T,t_1,i}
            \big)$\;
            $g_{\mathrm{static},t_2,i}\leftarrow
            2p_i\big(
            \tilde{f}_{S,t_2,i}-\tilde{f}_{T,t_2,i}
            \big)$\;

            $e_{\mathrm{trans},i}\leftarrow
            \big(
            \tilde{f}_{S,t_1,i}-\tilde{f}_{S,t_2,i}
            \big)
            -
            \big(
            \tilde{f}_{T,t_1,i}-\tilde{f}_{T,t_2,i}
            \big)$\;
            $g_{\mathrm{trans},i}\leftarrow
            2\lambda_{\mathrm{trans}}p_i e_{\mathrm{trans},i}$\;
        }

        \tcp{Step 5: Inject fused directions and update the student}
        treat $\Omega_{\mathrm{trans}}$, $p_i$, $\alpha_i$, and all centering statistics as stop-gradient quantities\;
        $\mathcal{L}_{\mathrm{Dynamic}}\leftarrow
        \sum_{i\in\Omega_{\mathrm{trans}}}
        \Big[
        f_{S,t_1,i}\cdot
        \operatorname{sg}\!\big(
        g_{\mathrm{static},t_1,i}+g_{\mathrm{trans},i}
        \big)
        +
        f_{S,t_2,i}\cdot
        \operatorname{sg}\!\big(
        g_{\mathrm{static},t_2,i}-g_{\mathrm{trans},i}
        \big)
        \Big]$\;
        update student model
        $\theta_S\leftarrow
        \theta_S-\eta\nabla_{\theta_S}
        \mathcal{L}_{\mathrm{Dynamic}}$\;
    }
    \KwRet{$\theta_S$}
\end{algorithm}
\paragraph{Training On D2F.}
When applied to the D2F architecture, the data preparation adheres to the block-wise partitioning and progressive noise scheduling of the baseline, with the network optimized through the BCSM mechanism.

\begin{itemize}
    \item \textit{Sample Construction and Noise Scheduling.} Following D2F, a clean text sequence $Y^0$ is divided into $N$ distinct blocks of size $k$. During the forward process, a monotonically increasing continuous noise schedule $\{t_1 < t_2 < ... < t_N\}$ is assigned to the $N$ blocks. Each block $Y_{B_i}$ is independently corrupted into $Y_{B_i}^{t_i}$ by replacing tokens with the \texttt{[MASK]} token with probability $t_i$. Consequently, earlier blocks retain more clean tokens while later blocks are heavily masked, simulating an autoregressive progression from known context to uncertain predictions.
    
    \item \textit{Block-Wise Causal Attention.} As established in the D2F framework, the student model utilizes a block-wise causal attention mask. This restricts the prediction of the $i$-th block to be conditioned strictly on the partially denoised preceding blocks and the current noisy block. This architecture forces the model to deduce future blocks from incomplete premises, enabling exact KV caching and inter-block parallel decoding during inference.
    
    \item \textit{BCSM Distillation Process.} Building upon this setup, we replace the original asymmetric KL divergence distillation with the proposed BCSM to prevent probability leakage. The teacher model operates with standard bidirectional full attention to provide global-view predictions, while the student model utilizes the aforementioned block-wise causal attention. Unnormalized logits ($f_T$ and $f_S$) are extracted at the masked positions, the Kneedle algorithm determines the dynamic active subspace, and the mean-centered variance identity computes the $\mathcal{O}(K)$ BCSM loss to update the student model.
\end{itemize}

\paragraph{Framework-Specific Hyperparameters.}
The preceding paragraphs describe the framework-specific sequence
construction, corruption process, attention pattern, and application
of the BCSM objective. To avoid repetition, we report here only
the explicitly configured model, optimization, and BCSM
hyperparameters. Parameters that are not separately configured,
as well as task-specific inference thresholds, are omitted.

Table~\ref{tab:bard_bcsm_hyperparameters} reports the settings
for the BARD-VL mask-diffusion experiments. Both the teacher and
student use the 4B backbone, but operate with block sizes $4$ and
$32$, respectively. We directly optimize the block-size-$32$
student under the supervision of the frozen block-size-$4$ teacher,
without progressive block merging or multi-stage distillation.
Table~\ref{tab:d2f_bcsm_hyperparameters} reports the corresponding
settings for D2F, for which the original parameter-efficient
optimization and block-wise training configurations are retained.


\begin{table*}[t]
\centering
\caption{\textbf{Explicit hyperparameters for BCSM distillation
under the BARD-VL mask-diffusion setting.}
Both teacher and student use the BARD-VL-4B backbone.
A frozen teacher with block size $4$ directly supervises a
student with block size $32$, without progressive block merging
or multi-stage distillation.}
\label{tab:bard_bcsm_hyperparameters}

\small
\setlength{\tabcolsep}{6pt}
\renewcommand{\arraystretch}{1.14}

\begin{tabularx}{\textwidth}{
    @{}
    >{\raggedright\arraybackslash}p{0.27\textwidth}
    >{\raggedright\arraybackslash}X
    @{}
}
\toprule
\textbf{Configuration}
&
\textbf{Setting}
\\

\midrule
\multicolumn{2}{@{}l}{\textbf{Model and distillation}} \\

Training variant
&
BARD-VL mask diffusion
\\

Teacher
&
Frozen BARD-VL-4B with block size $B_{\mathrm{T}}=4$
\\

Student
&
BARD-VL-4B with block size $B_{\mathrm{S}}=32$
\\

Distillation route
&
Direct single-stage cross-block distillation:
$4\mathrm{B}\,(B_{\mathrm{T}}=4)
\rightarrow
4\mathrm{B}\,(B_{\mathrm{S}}=32)$
\\

Trainable parameters
&
All student parameters
\\

\midrule
\multicolumn{2}{@{}l}{\textbf{Optimization}} \\

Optimizer
&
AdamW with $\beta_1=0.9$ and $\beta_2=0.95$
\\

Peak learning rate
&
$2\times10^{-5}$ for the language model and
$2\times10^{-6}$ for the vision tower
\\

Learning-rate scheduler
&
Linear warmup followed by cosine decay to $1\times10^{-6}$
\\

Warmup
&
$1{,}000$ optimization steps
\\

Global batch size
&
$256$
\\

Gradient clipping
&
Maximum gradient norm of $1.0$
\\

\midrule
\multicolumn{2}{@{}l}{\textbf{BCSM}} \\

Maximum search range $K_{\max}$
&
$1024$
\\

Transition coefficient $\lambda_{\mathrm{Trans}}$
&
$0.1$ for the transition-aware BCSM objective;
not used for static BCSM
\\

\bottomrule
\end{tabularx}
\end{table*}


\begin{table*}[t]
\centering
\caption{\textbf{Explicit hyperparameters for BCSM distillation
under the D2F setting.}
The original D2F backbone initialization, LoRA optimization,
sequence partitioning, and block-wise masking range are retained,
while its asymmetric KL objective is replaced by BCSM.}
\label{tab:d2f_bcsm_hyperparameters}

\small
\setlength{\tabcolsep}{6pt}
\renewcommand{\arraystretch}{1.14}

\begin{tabularx}{\textwidth}{
    @{}
    >{\raggedright\arraybackslash}p{0.27\textwidth}
    >{\raggedright\arraybackslash}X
    @{}
}
\toprule
\textbf{Configuration}
&
\textbf{Setting}
\\

\midrule
\multicolumn{2}{@{}l}{\textbf{Model and distillation}} \\

Backbones
&
LLaDA-Instruct-8B and Dream-Base-7B
\\

Teacher
&
Frozen pretrained diffusion language model with bidirectional
full attention
\\

Student initialization
&
Initialized from the corresponding pretrained teacher backbone
\\

Trainable parameters
&
Student LoRA parameters
\\

\midrule
\multicolumn{2}{@{}l}{\textbf{Optimization}} \\

Optimizer
&
AdamW
\\

Learning rate
&
Constant learning rate of $1\times10^{-5}$
\\

LoRA target modules
&
\texttt{q\_proj}, \texttt{k\_proj},
\texttt{v\_proj}, and \texttt{o\_proj}
\\

LoRA rank
&
$r=32$
\\

LoRA scaling factor
&
$\alpha=32$
\\

LoRA dropout
&
$0.1$
\\

\midrule
\multicolumn{2}{@{}l}{\textbf{Sequence and block configuration}} \\

Training sequence length
&
$512$ tokens
\\

Training block size
&
$k=16$
\\

Block-wise mask-ratio range
&
Monotonically increasing from $0.3$ to $0.7$
\\

\midrule
\multicolumn{2}{@{}l}{\textbf{BCSM}} \\

Maximum search range $K_{\max}$
&
$1024$
\\

Transition coefficient $\lambda_{\mathrm{Trans}}$
&
$0.1$ for the transition-aware BCSM objective;
not used for static BCSM
\\

\bottomrule
\end{tabularx}
\end{table*}

\subsection{Proof of the Variance Identity for Mean-Centered Factorization}
\label{appendix:variance_proof}

In Section~\ref{subsec:bcsm} of the main text, we introduced a
mean-centered formulation to reduce the active-set pairwise log-score
matching complexity from $\mathcal{O}(K^2)$ to $\mathcal{O}(K)$.
The theoretical foundation of this factorization is the weighted
variance identity.
Here, we provide the formal proof for
Eq.~\eqref{eq:variance_identity}.

\textbf{Lemma 1 (Weighted Variance Identity).}
\textit{Let $\Omega$ be a discrete state space and $D$ be a scalar
function over $\Omega$.
Given normalized nonnegative weights $\alpha$ such that
$\sum_{k\in\Omega}\alpha_k=1$, define the weighted mean
$\bar{D}=\sum_{k\in\Omega}\alpha_kD_k$.
Then,}
\begin{equation}
\frac{1}{2}
\sum_{i\in\Omega}
\sum_{j\in\Omega}
\alpha_i\alpha_j(D_i-D_j)^2
=
\sum_{i\in\Omega}
\alpha_i(D_i-\bar{D})^2.
\end{equation}

\begin{proof}
We begin by expanding the squared term on the left-hand side (LHS):
\begin{equation}
\mathrm{LHS}
=
\frac{1}{2}
\sum_{i\in\Omega}
\sum_{j\in\Omega}
\alpha_i\alpha_j
(D_i^2-2D_iD_j+D_j^2).
\end{equation}
Since the summation indices $i$ and $j$ are independent, we can
distribute the sums and decouple the terms:
\begin{equation}
\begin{split}
\mathrm{LHS}
=
\frac{1}{2}\Bigg[
&\left(\sum_{i\in\Omega}\alpha_iD_i^2\right)
 \left(\sum_{j\in\Omega}\alpha_j\right)\\
&-2\left(\sum_{i\in\Omega}\alpha_iD_i\right)
   \left(\sum_{j\in\Omega}\alpha_jD_j\right)\\
&+\left(\sum_{i\in\Omega}\alpha_i\right)
  \left(\sum_{j\in\Omega}\alpha_jD_j^2\right)
\Bigg].
\end{split}
\end{equation}
Using
$\sum_{k\in\Omega}\alpha_k=1$ and
$\sum_{k\in\Omega}\alpha_kD_k=\bar{D}$, the expression becomes
\begin{equation}
\begin{aligned}
\mathrm{LHS}
&=
\frac{1}{2}
\Bigg[
\left(\sum_{i\in\Omega}\alpha_iD_i^2\right)
-2\bar{D}^2
+\left(\sum_{j\in\Omega}\alpha_jD_j^2\right)
\Bigg]\\
&=
\sum_{i\in\Omega}\alpha_iD_i^2-\bar{D}^2.
\end{aligned}
\end{equation}

Next, we expand the right-hand side (RHS):
\begin{equation}
\mathrm{RHS}
=
\sum_{i\in\Omega}
\alpha_i(D_i-\bar{D})^2
=
\sum_{i\in\Omega}
\alpha_i(D_i^2-2D_i\bar{D}+\bar{D}^2).
\end{equation}
Distributing the summation yields
\begin{equation}
\begin{aligned}
\mathrm{RHS}
&=
\sum_{i\in\Omega}\alpha_iD_i^2
-2\bar{D}
\left(\sum_{i\in\Omega}\alpha_iD_i\right)
+\bar{D}^2
\left(\sum_{i\in\Omega}\alpha_i\right)\\
&=
\sum_{i\in\Omega}\alpha_iD_i^2
-2\bar{D}^2+\bar{D}^2\\
&=
\sum_{i\in\Omega}\alpha_iD_i^2-\bar{D}^2.
\end{aligned}
\end{equation}
Since $\mathrm{LHS}=\mathrm{RHS}$, the identity follows.
\end{proof}

\textbf{Connection to the Dynamic Active Subspace.}
In BCSM, the student probabilities are computed over the full
vocabulary as
$p_i=\operatorname{Softmax}_{\mathcal{V}}(f_S)_i$.
For the active subspace
$\Omega\subseteq\mathcal{V}$ selected using the Kneedle-based boundary
criterion, define its retained probability mass and normalized
centering weights as
\begin{equation}
W_{\Omega}
=
\sum_{k\in\Omega}p_k,
\qquad
\alpha_i
=
\frac{p_i}{W_{\Omega}}.
\end{equation}
Here, $0<W_{\Omega}\leq1$, and the discarded probability mass is
$1-W_{\Omega}$.
Applying Lemma~1 with $\alpha_i=p_i/W_{\Omega}$ and multiplying both
sides by $W_{\Omega}$ gives
\begin{equation}
\frac{1}{2W_{\Omega}}
\sum_{i\in\Omega}
\sum_{j\in\Omega}
p_ip_j(D_i-D_j)^2
=
\sum_{i\in\Omega}
p_i(D_i-\bar{D})^2,
\label{eq:variance_identity_appendix}
\end{equation}
where
\begin{equation}
\bar{D}
=
\sum_{k\in\Omega}\alpha_kD_k.
\end{equation}
Setting
$D_i=f_{S,i}-f_{T,i}$ and centering the teacher and student logits
using the same weights $\alpha_i$ yields
\begin{equation}
\bar{D}
=
\bar{f}_S-\bar{f}_T,
\qquad
D_i-\bar{D}
=
\tilde{f}_{S,i}-\tilde{f}_{T,i}.
\end{equation}
Therefore,
\begin{equation}
\begin{aligned}
&\frac{1}{2W_{\Omega}}
\sum_{i\in\Omega}
\sum_{j\in\Omega}
p_i p_j
\left[
(f_{S,i}-f_{S,j})
-
(f_{T,i}-f_{T,j})
\right]^2 \\
&\qquad =
\sum_{i\in\Omega}
p_i
\left(
\tilde{f}_{S,i}-\tilde{f}_{T,i}
\right)^2.
\end{aligned}
\end{equation}
The normalization by $W_{\Omega}$ is used only to define the
centering weights $\alpha_i$, while the pointwise objective retains
the full-vocabulary probabilities $p_i$ as its outer weights.
Thus, the factorization is exact for any fixed active subspace with
$W_{\Omega}>0$, reducing its pairwise matching complexity from
$\mathcal{O}(K^2)$ to $\mathcal{O}(K)$.
During surrogate gradient injection, $\Omega$, $p_i$, $\alpha_i$, and
the derived centering statistics are treated as stop-gradient
quantities.

\begin{figure*}[t]
    \centering
    \includegraphics[width=\textwidth]{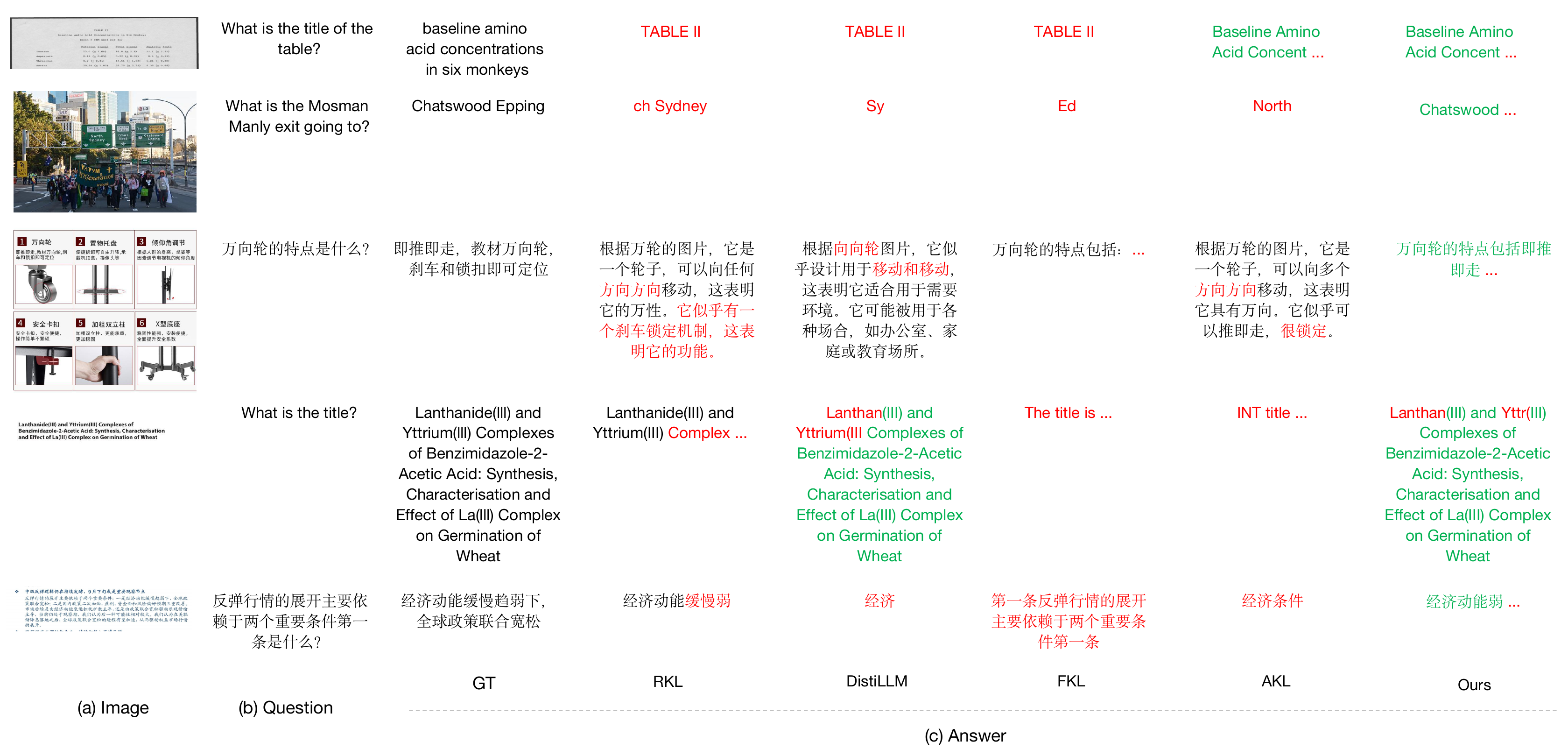} 
    \caption{Qualitative comparison of cases where all methods fail to recover the exact reference. In these examples, KL-based methods produce fragmented tokens or repetitions, whereas BCSM retains longer recognizable spans and more complete local structures. The comparison reports an empirical property of the generated outputs rather than explicit cross-position dependency modeling.}
    \label{fig:divergent_generations}
\end{figure*}

\subsection{Empirical Analysis of Locally Inconsistent Generations in Multi-modal Understanding}
\label{app:probability_leakage}

As mathematically analyzed in Section \ref{subsec:kl_to_sm}, the application of locally factorized KL divergence in parallel decoding inherently leads to multi-modality collapse and sequence-level incoherence. 
Figure \ref{fig:probability_leakage_vis} provides direct empirical evidence of this vulnerability in multi-modal understanding tasks. 
When employing standard KL-based distillation methods ($e.g.$, RKL, DistiLLM, FKL, AKL), the generated responses systematically exhibit degenerative artifacts. 
Because the token-independent KL objective forces the model to match marginal probabilities at each position without explicit inter-token sampling dependencies, probability mass is misallocated to invalid semantic combinations. 
This phenomenon empirically surfaces in three primary forms: spelling errors ($e.g.$, generating ``F-PRR'' or ``nofofcsereGtr''), improper structural compositions yielding incorrect semantics ($e.g.$, generating mixed concepts like ``Classic and British Cars 2016 Calendar'' or \includegraphics[height=0.9em]{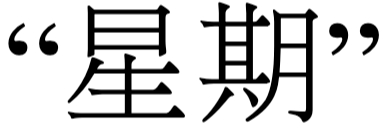}), and meaningless token repetitions ($e.g.$, ``4GBGB+64GB+64GBGB'', \includegraphics[height=0.9em]{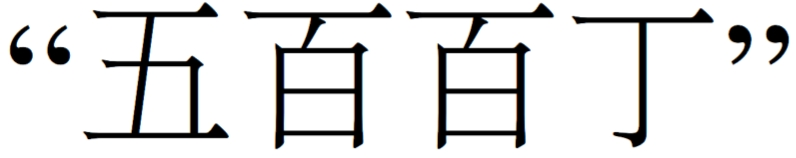}). 
By transitioning to score matching and matching the unnormalized relative logit geometry, our proposed approach implicitly distills the contextual dependencies captured by the teacher, effectively eliminating these locally inconsistent generation artifacts and ensuring semantically valid predictions.

\subsection{Semantic Stability of Score Matching in Divergent Generations}
\label{app:semantic_stability}

In addition to the analysis of exact match accuracy, it is necessary to examine instances where all decoding methods deviate from the exact ground truth. 
Figure \ref{fig:divergent_generations} presents such cases, providing further empirical evidence for the structural advantages of the proposed score matching approach theoretically discussed in Section \ref{subsec:kl_to_sm}. 
When standard KL distillation models fail to predict the exact target, they frequently exhibit severe structural degradation, yielding fragmented outputs ($e.g.$, ``Sy'', ``Ed'') or token repetitions ($e.g.$, \includegraphics[height=0.9em]{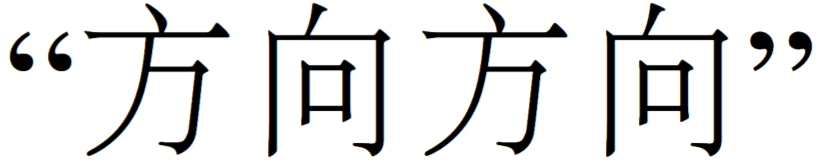}). 
This degradation occurs because the localized conditional independence assumption lacks a structured fallback mechanism when the primary mode is missed. 

Conversely, the proposed approach maintains structural integrity and semantic completeness even during divergent predictions. 
By preserving the teacher's relative energy topology via score matching, the model establishes a topological relationship for semantic transitions between tokens. 
Consequently, even when the exact ground truth is not fully recovered, the generated sequence remains constrained to a valid semantic manifold. 
For example, our method generates coherent partial semantic structures such as ``Chatswood ...'' and ``Baseline Amino Acid Concent'' instead of disjointed characters. 
Similarly, for Chinese queries, it generates coherent continuations ($e.g.$, \includegraphics[height=0.98em]{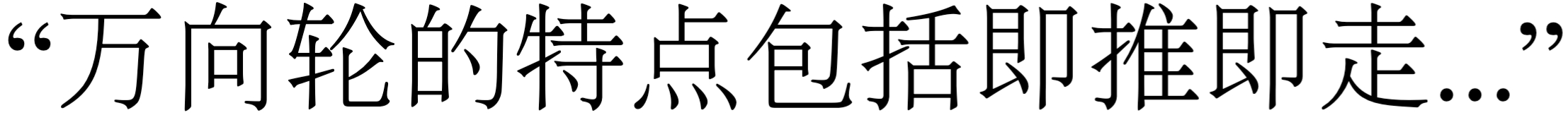} and \includegraphics[height=0.98em]{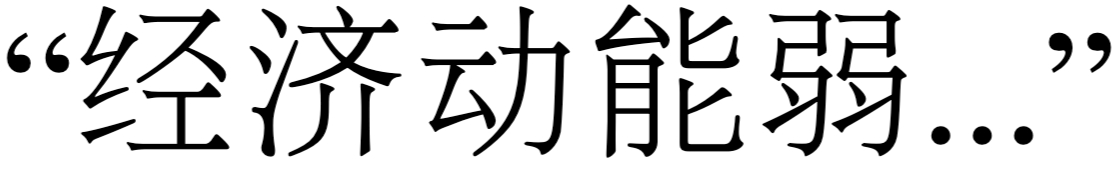}) rather than meaningless semantic loops.
These empirical observations validate that establishing relative energy topologies at each position via score matching preserves semantic stability, ensuring that even suboptimal responses maintain logical coherence.

\end{document}